\documentclass[sigconf]{acmart}
\AtBeginDocument{%
  \providecommand\BibTeX{{%
    \normalfont B\kern-0.5em{\scshape i\kern-0.25em b}\kern-0.8em\TeX}}}

\setcopyright{acmcopyright}
\copyrightyear{2021}
\acmYear{2021}
\setcopyright{rightsretained}
\acmConference[ACM CHIL '21]{ACM Conference on Health, Inference, and Learning}{April 8--10, 2021}{Virtual Event, USA}
\acmBooktitle{ACM Conference on Health, Inference, and Learning (ACM CHIL '21), April 8--10, 2021, Virtual Event, USA}\acmDOI{10.1145/3450439.3451878}
\acmISBN{978-1-4503-8359-2/21/04}

\usepackage{enumitem}
\usepackage[title]{appendix}
\usepackage{graphics}
\usepackage{caption}
\usepackage{subcaption}
\usepackage[linesnumbered,ruled,vlined]{algorithm2e}
\usepackage{multirow}
\usepackage{bm}
\usepackage{amsmath}
\usepackage{placeins}

\DeclareMathOperator*{\argmax}{\arg\!\max}

\begin{document}

\title{An Empirical Framework for Domain Generalization in Clinical Settings}

\author{Haoran Zhang}
\email{haoran@cs.toronto.edu}
\affiliation{%
  \institution{University of Toronto}
  \institution{Vector Institute\\[3.0ex]}
  \country{}
}

\author{Natalie Dullerud}
\email{dullerud@cs.toronto.edu}
\affiliation{%
  \institution{University of Toronto}
  \institution{Vector Institute}
  \country{}
}

\author{Laleh Seyyed-Kalantari}
\email{laleh@cs.toronto.edu}
\affiliation{%
  \institution{University of Toronto}
  \institution{Vector Institute}
  \country{}
}

\author{Quaid Morris}
\email{morrisq@mskcc.org}
\affiliation{%
  \institution{Memorial Sloan Kettering Cancer Center\\[3.0ex]}
  \country{}
}

\author{Shalmali Joshi}
\email{shalmali@seas.harvard.edu}
\affiliation{%
  \institution{Harvard University}
  \country{}
}

\author{Marzyeh Ghassemi}
\email{marzyeh@cs.toronto.edu}
\affiliation{%
  \institution{University of Toronto}
  \institution{Vector Institute}
  \country{}
}

\renewcommand{\shortauthors}{Zhang et al.}

\begin{abstract}
  Clinical machine learning models experience significantly degraded performance in datasets not seen during training, e.g., new hospitals or populations. Recent developments in domain generalization offer a promising solution to this problem by creating models that learn invariances across environments. In this work, we benchmark the performance of eight domain generalization methods on multi-site clinical time series and medical imaging data. We introduce a framework to induce synthetic but realistic domain shifts and sampling bias to stress-test these methods over existing non-healthcare benchmarks. 
  We find that current domain generalization methods do not consistently achieve significant gains in out-of-distribution performance over empirical risk minimization on real-world medical imaging data, in line with prior work on general imaging datasets. However, a subset of realistic induced-shift scenarios in clinical time series data do exhibit limited performance gains. We characterize these scenarios in detail, and recommend best practices for domain generalization in the clinical setting.
\end{abstract}



\maketitle

\section{Introduction}
As machine learning models become more prevalent in clinical settings, it is important to consider how well models can generalize to environments external to their training environment~\cite{kellyKeyChallengesDelivering2019, subbaswamyDevelopmentDeploymentDataset2020, challenArtificialIntelligenceBias2019, castroCausalityMattersMedical2020}. Current large-scale clinical machine learning models often utilize data from a single site in urban population centers, such as the Beth Israel Deaconess Medical Center in Boston for the MIMIC-III dataset~\cite{johnson2016mimic}. If models trained on these datasets are deployed in other regions or countries, it is important that their performance degradation is minimal.

Prior work has found significant decreases in model performance under the presence of cross-institutional domain shift, in the chest X-ray~\cite{zechVariableGeneralizationPerformance2018, poochCanWeTrust2020, cohenLimitsCrossdomainGeneralization2020}, MRI~\cite{martenssonReliabilityDeepLearning2020, albadawyDeepLearningSegmentation2018}, and pathology~\cite{stackeCloserLookDomain2019, stackeMeasuringDomainShift2020, thagaardCanYouTrust2020} settings. Temporal domain shifts have also been found to reduce performance in clinical machine learning models~\cite{nestorFeatureRobustnessNonstationary2019}. Recent developments in domain generalization present a way to combat this problem by learning models that are invariant across environments while ignoring environment-specific spurious correlations~\cite{arjovskyInvariantRiskMinimization2019}.

In this work, we focus on the domain generalization learning setup, where a model is learnt on data from multiple training environments, e.g., hospitals, labs, or regions, and evaluated directly on an unseen test environment without further fine-tuning~\cite{gulrajaniSearchLostDomain2020}. In our setting, no data from the test environment is accessible to the model during training. 

There are several methods that have been developed for domain generalization. The naive baseline is to use empirical risk minimization (ERM) to learn a single model on pooled data across all training environments. Another approach is invariant causal prediction, which assumes the existence of a shared causal graph across all environments, and seeks to discover a subset of invariant features using conditional independence tests~\cite{petersCausalInferenceUsing2015,heinze-demlInvariantCausalPrediction2018}. Recent extensions of this work relax many of its assumptions and are computationally feasible for large datasets~\cite{arjovskyInvariantRiskMinimization2019, ahuja2020invariant}. Other methods attempt to learn a representation that has the same distribution across the training environments~\cite{ganin2016domain, li2018domain, moyer2018invariant, deng2020representation}, e.g., with an adversary, or attempt meta-learning from the assumed meta-distribution where all environments are drawn~\cite{li2018learning, dou2019domain}. The computer vision literature has also created methods that rely on data augmentation techniques and auxiliary tasks which are specific to the image domain~\cite{carlucci2019domain, bentonLearningInvariancesNeural2020}. 

In this work, we focus on domain generalization methods which are data modality agnostic, i.e., can be applied to tabular, time series, or image data alike. Domain generalization methods in the literature have been largely benchmarked on datasets where spurious correlations are introduced in a contrived manner, such as Colored MNIST~\cite{arjovskyInvariantRiskMinimization2019} or Colored Fashion MNIST~\cite{ahuja2020invariant}. More realistic recent evaluations have demonstrated that no domain generalization algorithm significantly outperforms ERM on standard image classification datasets~\cite{gulrajaniSearchLostDomain2020}. Similarly, \citet{kohWILDSBenchmarkIntheWild2020a} found that two domain generalization methods often actually perform significantly worse than ERM on seven real-world datasets spanning text, image, and graph modalities. 

We evaluate the performance of eight algorithms on domain generalization in \textit{clinical} time series data from intensive care units (ICUs) across four regions~\cite{pollard2018eicu} and chest x-ray imaging data from four sites~\cite{johnson_mimic-cxr:_2019, irvin_chexpert:_2019, wang_chestx-ray8:_2017, bustos_padchest:_2019}. The clinical setting presents a realistic domain for benchmarking methods that might be trained in one site, but deployed in another. We also manually introduce realistic sampling bias in the data to test the limits of these methods in sites with further shift. We present these clinical confounding and sampling bias scenarios as a general empirical framework to stress-test generalization methods. Our main contributions are the following:
\begin{itemize}[topsep=2pt]
    \item We show that state of the art domain generalizations do not consistently perform significantly better than ERM on real-world clinical imaging data. This is consistent with results from prior work on general benchmarking datasets~\cite{kohWILDSBenchmarkIntheWild2020a, gulrajaniSearchLostDomain2020}.
    \item We introduce a framework which generates plausible augmented versions of clinical datasets with domain shift. While there are realistic clinical scenarios where domain generalization perform marginally better than ERM, these improvements only manifest when the strength of the spurious correlation is strong.
    \item We find, in the case of subsampled datasets with varying label prevalence between genders, that domain generalization methods are not able to learn fairer models than ERM while maintaining overall model performance.
    \item We publicly release the code and framework to reproduce our data and results\footnote{\href{https://github.com/MLforHealth/ClinicalDG}{\texttt{https://github.com/MLforHealth/ClinicalDG}}}, based on a modified version of the DomainBed~\cite{gulrajaniSearchLostDomain2020} platform.
\end{itemize}

We hope this framework will be used as a realistic clinical generalization scenario against which domain generalization methods can be benchmarked.

\section{Related Works}

\subsection{Domain Generalization Methods}
\label{subsec:domaingenmethods}
In the domain generalization learning setup, we are given labelled data from multiple training environments, and seek to learn a model whose performance generalizes to unseen test environments. Approaches based on causality stemmed from the Invariant Causal Prediction (ICP) method proposed by \citet{petersCausalInferenceUsing2015}, which assumes the existence of a causal graph and uses conditional independence tests to find a set of invariant features. Followup work include extensions to non-linear models~\cite{heinze-demlInvariantCausalPrediction2018} and the use of anchor variables~\cite{rothenhauslerAnchorRegressionHeterogeneous2019}. However, finding this invariant feature set involves a combinatorial search over the feature space, and these conditional independence tests often make many distributional assumptions. 

Domain generalization from robust optimization \cite{ben2009robust} seeks to minimize the worst-case error in the training environments. \citet{kruegerOutofDistributionGeneralizationRisk2020} introduced the principle of risk extrapolation, which is a generalized form of robust optimization. \citet{xieRiskVariancePenalization2020} derived a slightly altered risk extrapolation loss function and linked distributional robustness with causality. Methods like GroupDRO \cite{sagawa2019distributionally}, conventionally used in the subpopulation shift setting, has also been tested for domain generalization \cite{gulrajaniSearchLostDomain2020}.

Another approach to domain generalization aims to remove all environment information from a latent representation, or, alternatively, learn an encoder such that all environments have the same latent distribution. This can be accomplished with an adversarial network~\cite{ganin2016domain, deng2020representation}, the Maximum Mean Discrepency (MMD) loss~\cite{li2018domain}, or by directly minimizing mutual information~\cite{moyer2018invariant}. Methods based on low-rank decomposition have also been proposed. These methods seek to learn a component that is common among all environments, and a component that is specific to each training domain. The common component is then used for out-of-distribution (OOD) generalization~\cite{piratlaEfficientDomainGeneralization2020, liDeeperBroaderArtier2017}.

Several methods have been proposed specific to the image domain. \citet{zhangWhenUnseenDomain2019} proposed a data augmentation based approach where a series of stacked transformations are applied. \citet{carlucci2019domain} proposed an auxiliary task for neural network training where the network learns to solve a jigsaw puzzle consisting of shuffled patches of an image. \citet{bentonLearningInvariancesNeural2020} introduced a method where the model automatically learns invariant affine augmentations from the training data. \citet{hendrycks2020many} proposed an image augmentation method involving applying randomly sampled operations to the weights and activations of an image autoencoder, though it could potentially be applicable to other modalities as well. In this work, we focus only on methods that are modality agnostic.

The invariant risk minimization (IRM) method proposed by \citet{arjovskyInvariantRiskMinimization2019} frames domain generalization as a bi-level optimization problem. In addition to alleviating the distributional assumptions of ICP, their optimization problem can be simplified to a loss function compatible with gradient descent that can easily be applied to large datasets. \citet{ahuja2020invariant} proposed an alternate method for solving the same bilevel optimization problem by finding the Nash equilibrium of an ensemble game.

\subsection{Model Transferrability in Medical Settings}
Access to large annotated datasets to train deep neural networks across multiple sites is not always feasible in clinical settings. Transfer learning ~\cite{Pan2009} addresses this by using a model pretrained on a large-scale dataset and fine-tuning it to the downstream task. This method has been commonly used in designing medical image classifiers~\cite{irvin_chexpert:_2019,rajpurkar_deep_2018,wang_chestx-ray8:_2017, CheXclusion_2020, akbarian2020evaluating}. In these settings, the deep neural network is initialized with a pretrained model (for example, trained on ImageNet ~\cite{imagenet_cvpr09}) and then are finetuned on downstream medical images. Transfer learning has been shown to be effective at increasing model performance in  chest X-ray classifiers~\cite{rajpurkar_deep_2018, CheXclusion_2020}, though there are cases where a model trained from scratch can perform just as well~\cite{raghu_transfusion:_2019}.

In the transfer learning framework, we are given labelled data for the target domain. A related framework is unsupervised domain adaptation, where we are only given unlabelled data for the target domain. Unsupervised domain adaptation has also been applied to medical imaging~\cite{perone2019unsupervised, zhang2020collaborative, dong2018unsupervised}. 
Our benchmark focuses on the domain generalization setting, where only labeled data from multiple training environments are available, and the goal is to be able to generalize to all unseen test domains.

There have been a limited number of papers which apply domain generalization methods to health data. In their WILDS benchmark, \citet{kohWILDSBenchmarkIntheWild2020a} tested two domain generalization methods on the Camelyon17 dataset for tumor identification~\cite{bandi2018detection}, finding that they both performed worse than ERM by more than $10\%$ accuracy. \citet{ghimireLearningInvariantFeature2020} benchmarked the performance of the IRM Games method~\cite{ahuja2020invariant} on pneumonia detection in four chest X-ray environments, finding that it gave marginal improvements to OOD performance. \citet{bellot2020generalization} also test their proposed method on pneumonia detection using chest X-ray datasets from two hospitals. However, as there is a significant overlap between the training and test domains in their experimental setup, it would be better suited as a subpopulation shift problem \cite{kohWILDSBenchmarkIntheWild2020a} rather than a domain generalization one.

\subsection{Domain Generalization and Fairness}
Fairness criteria have grown in popularity in recent years due to the increasing use of machine learning models in settings such as healthcare~\cite{tomavsev2019clinically,rajkomar2018scalable,beam2018big,wu2019modeling, chen2020ethical, pfohl2021empirical}, where poor performance of models on certain subgroups can lead to significant harm. Common group fairness metrics, such as statistic parity, equalized odds, and equality of opportunity consider fairness through various independence definitions, typically between the random variables of the true label, predicted label and protected attribute (attribute determining subgroups)~\cite{hardt2016equality}. Many group fairness objectives focus on minimizing the worst-case performance or the gap in performance according to certain metrics (such as parity, recall, etc.) across subgroups~\cite{kearns2019subgroupfairness}.

Domain generalization methods, as described in Section~\ref{subsec:domaingenmethods}, similarly aim to minimize the worst-case risk across all possible environments. State-of-the-art algorithms, such as GroupDRO, have arguably been motivated by improvement to both group fairness and generalization performance~\cite{sagawa2019distributionally, deng2020representation}. There has been some recent literature investigating the relationship between domain generalization and fairness~\cite{creager2020exchanging, adragna2020fairness, deng2020representation}, and some analysis of group and individual fairness constraints on generalization ability~\cite{sharifi2019individualfairness, cotter2019fairness}. 

Within the IRM objective, \citet{creager2020exchanging} improves worst-case performance without access to protected group labels in order to develop a generalization method for settings in which the domain labels are not provided. This paper also demonstrated that the IRM objective can be framed to directly optimize group sufficiency if the protected attribute label is taken as the environment variable. \citet{adragna2020fairness} provided empirical results for the gains IRM offers over ERM in terms of fairness guarantees through comparing the ability of both objectives to be invariant to spurious correlations between comment toxicity and particular demographic groups in internet comment datasets. 

In this work, we add to existing empirical results linking domain generalization and fairness to investigate this relationship in a \textit{clinical} context.

\subsection{Domain Generalization Benchmarks}
The large majority of state-of-the-art domain generalization methods are tested on variants of MNIST (such as Colored MNIST) where a spurious correlation (such as a correlation between the channel and the label) are introduced synthetically~\cite{arjovskyInvariantRiskMinimization2019, paceLearningDiverseRepresentations2020, koyamaOutofDistributionGeneralizationMaximal2020}. \citet{choeEmpiricalStudyInvariant2020} proposed Extended Colored MNIST -- a version of Colored MNIST with varying data generation parameters. They benchmark the performance of IRM and ERM on this dataset, along with a sentiment analysis dataset where punctuation is manually confound with the label.

Two large-scale domain generalization benchmarks have been proposed. \citet{gulrajaniSearchLostDomain2020} proposed the DomainBed platform, which tests 15 methods on seven image benchmark datasets classically used for domain adaptation. One example is the PACS dataset~\cite{liDeeperBroaderArtier2017}, where the environments consist of photo, artistic, cartoon, or sketch renditions of objects. Though these datasets are much more realistic than Colored MNIST, they still have limited real-world utility. \citet{gulrajaniSearchLostDomain2020} found that domain generalization methods do not significantly out-perform ERM consistently.
\citet{kohWILDSBenchmarkIntheWild2020a} proposed the WILDS benchmark, which consists of seven real-world datasets spanning a variety of domains, including satellite imagery, cancer pathology, molecular graphs, and sentiment analysis. They tested two domain generalization methods -- IRM~\cite{arjovskyInvariantRiskMinimization2019} and DeepCORAL~\cite{sun2016deep} -- and found that neither of the methods improve over ERM performance on \textit{any} of the datasets.

In this work, we benchmark the performance of eight domain generalization methods on two real-world clinical datasets. In addition to the base datasets, we propose a framework for augmenting clinical datasets via synthetic domain shifts and sampling bias. We hope that this framework will bridge the gap between the state-of-the-art performance that domain generalization methods have shown on the contrived Colored MNIST dataset, and their poor performance on real-world datasets as demonstrated by the two other benchmarks.

\section{Methods}

\begin{figure*}[!h]
\centering
\includegraphics[width = 1.0\textwidth]{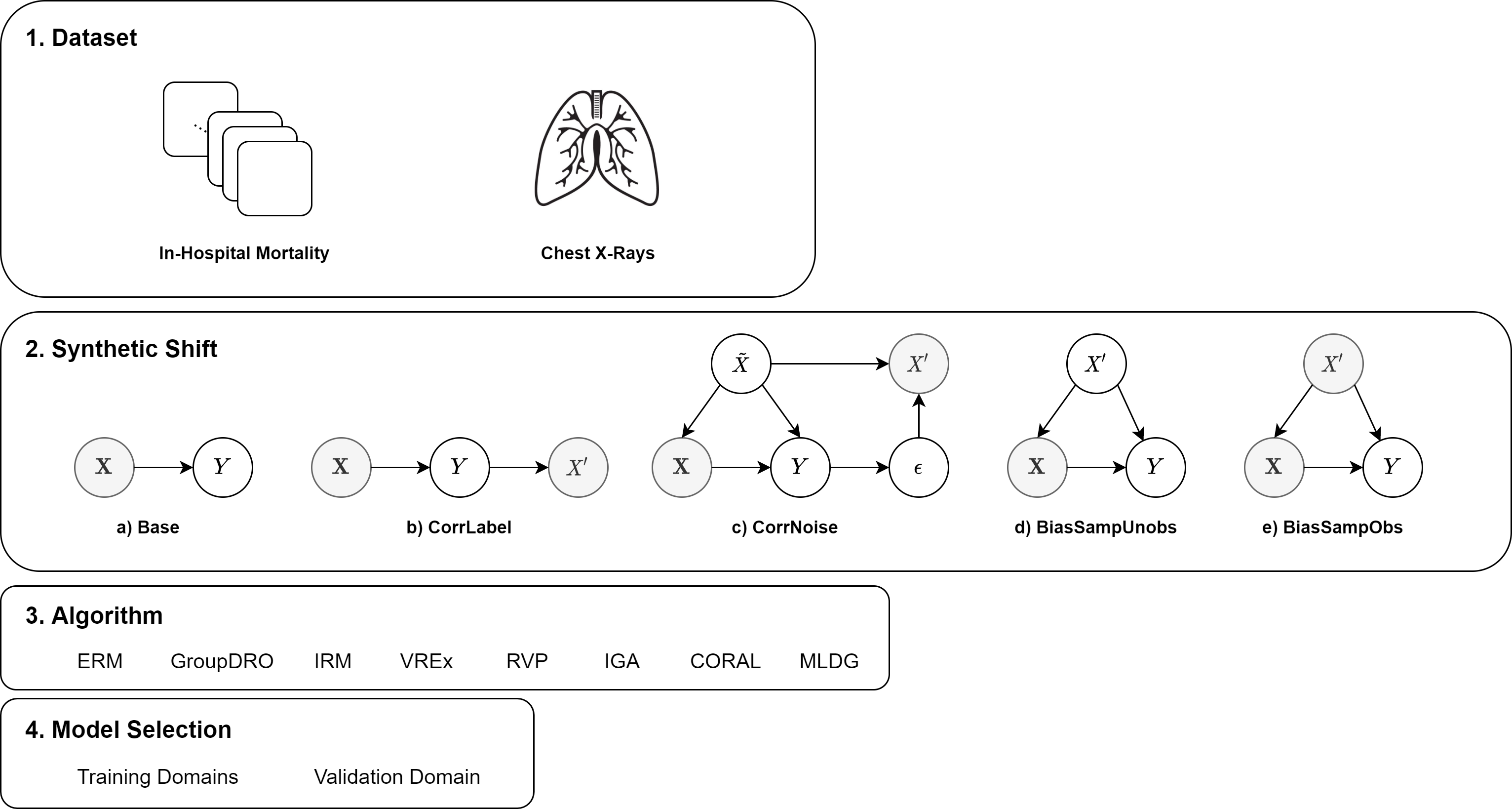}
\caption{Procedure for conducting domain generalization experiments. 1. We select a dataset consisting of multiple environments. 2. We choose a synthetic shift. Causal graphs are shown for \textbf{(a)} the base dataset; \textbf{(b)} addition of the corrupted label ($X'$) as a feature; \textbf{(c)} additional of noise $\epsilon$ that is correlated with the label to a feature $\tilde{X}$ to create a new feature $X'$; \textbf{(d)} subsampling based on a binary feature $X'$, where $X'$ is unobserved, and \textbf{(e)} subsampling based on a binary feature $X'$, where $X'$ is observed. Multi-dimensional random variables are shown in bold. Shaded nodes denote variables that are observed by the model. 3. We select a domain generalization algorithm. 4. We choose a strategy to conduct model selection.}
\label{fig:flow}
\end{figure*}

In the domain generalization setup, we are given labelled data $\{(x_i^e, y_i^e)\}_{i=1}^n$, from multiple training environments $e \in \mathcal{E}_{tr}$, as well as a risk function $R^e(f) = \mathbb{E}_{X^e, Y^e}[\ell(f(X^e), Y^e)]$. The goal is to learn a predictor $f: X \to Y$ that minimizes the worst-case risk across all possible environments $R^{OOD}(f) = \max_{e \in \mathcal{E}_{all}} R^e(f)$. In practice, we typically evaluate the performance of a domain generalization method by evaluating the risk of its learnt predictor on some unseen test environment $R^{e_{test}}(f)$.

\subsection{Domain Generalization Algorithms}
We test the performance of the following eight algorithms:
\begin{itemize}
    \item Empirical Risk Minimization (\textbf{ERM},~\cite{vapnik1992principles}) minimizes loss over pooled data across all training environments.
    \item Group Distributionally Robust Optimization (\textbf{GroupDRO},~\cite{sagawa2019distributionally}) minimizes the loss of the worst-case training environment. 
    \item Invariant Risk Minimization (\textbf{IRM},~\cite{arjovskyInvariantRiskMinimization2019}) learns a predictor that is invariant across training environments by optimizing the data representation such that all domains have the same downstream classifier.
    \item Variance Risk Extrapolation (\textbf{VREx},~\cite{kruegerOutofDistributionGeneralizationRisk2020}) minimizes the training risks along with the variance of the training risks across environments.
    \item Risk Variance Penalization (\textbf{RVP},~\cite{xieRiskVariancePenalization2020}) minimizes the training risks along with the standard deviation of the training risks across environments.
    \item Maximal Invariant Predictor by Inner-environmental Gradient Alignment (\textbf{IGA},~\cite{koyamaOutofDistributionGeneralizationMaximal2020}) learns the optimal classifier such that the label is independent of the environment index given the data representation.
    \item Deep Correlation Alignment for Deep Domain Adaptation (\textbf{CORAL},~\cite{sun2016deep}) aligns the mean and covariance of latent distributions across domains.
    \item Meta-Learning for Domain Generalization (\textbf{MLDG},~\cite{li2018learning}) adapts the model-agnostic meta-learning method~\cite{finn2017model} to the domain generalization setting.
\end{itemize}

We also report the performance of two ``oracles'', corresponding to models that \textit{do} have access to the test environment during training:
\begin{itemize}
    \item \textbf{OracleID}: train an ERM classifier only on the training split of the test environment. Note that this is not invariant model, as it would learn spurious correlations that exist on the test domain.
    \item \textbf{OracleMerged}: train an ERM classifier on the pooled training splits across all environments.
\end{itemize}

The difference in performance between the oracles and the ERM model is a proxy measure of how distinct the test environment is from the training environments. The performance of the oracle models is also an informal upper bound for the performance that any of the eight algorithms can hope to achieve. 

\subsection{Model Selection Strategies}
Model selection is a crucial part of a domain generalization experiment~\cite{gulrajaniSearchLostDomain2020}. It is not realistic to assume that the test environment is available for model selection, i.e., during hyperparameter tuning or early stopping, as is done in Colored MNIST; \citet{gulrajaniSearchLostDomain2020} observed that most of the performance gains on Colored MNIST by domain generalization methods vanish when the test environment is not used for model selection.

We consider two potential model selection methods for all experiments to critically evaluate the impact of the model selection policy on generalization performance:
\begin{itemize}
    \item \textbf{Selection by Training Domain}: We split the data for each training environment into training, validation, and test sets. We use the validation sets pooled across all training environments for model selection. This model selection method does not require any data external to the training environments, but it is unclear that the training domain validation sets would be a good measure of OOD performance.
    
    \item \textbf{Selection by Validation Domain}: We designate a specific environment as the validation environment. The data from the validation environment is used only for model selection. In later manual augmentation experiments, we assign the validation environment to have an intermediate level of spurious correlation, between the training environments and the test environment. This simulates the case where limited information is available from an environment closer to the deployment target.
    
\end{itemize}

\section{Synthetic Domain Shift Framework}
\label{sec:framework}

We experiment with five types of synthetic domain shifts, shown in Figure~\ref{fig:flow}. the unmodified dataset (\texttt{Base}), a noise corrupted label (\texttt{CorrLabel}), a feature-correlated corrupted label (\texttt{CorrNoise}), and biased subsampling (\texttt{BiasSampUnobs} and \texttt{BiasSampObs}).

\subsection{Unmodified Dataset (\texttt{Base})}
\label{subsec:no_confounding}
This corresponds to the \texttt{Base} graph shown in Figure~\ref{fig:flow}. For simplicity, we merge all features into a single node $\mathbf{X}$. However, complex causal relationships exist between the features and the label -- some of the features may be invariant, and some of which may be spuriously correlated with the label.

\subsection{Corrupted Label as Feature (\texttt{CorrLabel})}
We create a new binary feature $X'$ by flipping the target $Y$ with a certain environment specific probability $p_e$. We append this feature to the dataset and treat it as a static feature during modelling. The causal graph for this augmentation is shown in Figure~\ref{fig:flow}. 

We fix the flip probability for the validation and test environment to $p_{val} = 0.5$ and $p_{test} = 0.9$. For the training environments, we use $(p_{e1}, p_{e2}, p_{e3}) = (\beta - \delta, \beta, \beta + \delta)$, where $\beta$ is the mean probability between the three environments, and $\delta$ is the distance between each environment. As IRM requires that the training environments are diverse enough to learn invariances~\cite{arjovskyInvariantRiskMinimization2019}, some distance between the training environments is required. We fix $\delta = 0.1$, and vary $\beta \in \{0.1, 0.3, 0.5\}$.

Here, the goal of domain generalization is to learn a model that completely ignores $X'$, as its correlation with the label $p(Y|X')$ is varying in the training environments, and is flipped for the test environment. In the medical setting, this would represent a scenario where a strong spurious correlation exists in one environment, and is not generalizable to external environments. For this augmentation, we also include the performance of the Unaugmented ERM model (\textbf{ERM Unaug}) for reference, which is the performance of the ERM model from Section \ref{subsec:no_confounding}. This is the performance of a model that ignores the spurious correlation completely.

\subsection{Correlated Noise (\texttt{CorrNoise})}
We modify an existing continuous feature $\tilde{X}$ to create $X'$ by adding Gaussian noise that is correlated with both the label and the environment. 
A practical scenario reflecting this setting is one where sicker patients exhibit more extreme values of a feature in some environments. We sample $\epsilon \sim \mathcal{N}(\lambda_e y, \sigma^2)$, where $\lambda_e$ is an environment specific hyperparameter, and $y \in \{-1, +1\}$ is the label. We define $X' = \tilde{X} + \epsilon$, and use $X'$ in place of $\tilde{X}$ as a feature in our model. This corresponds to the \texttt{CorrNoise} causal graph in Figure~\ref{fig:flow}. 

We set $\lambda_{val} = 0.0$ and $\lambda_{test} = -1.0$, and we fix $\sigma = 0.5$. For the training environments, we set $(\lambda_{e1}, \lambda_{e2}, \lambda_{e3})$ as $(\beta - \delta, \beta, \beta + \delta)$ respectively. We vary $\beta \in \{1.0, 2.0\}$ and $\delta \in \{0.1, 0.5\}$.

Similar to \texttt{CorrLabel}, the goal is to learn a model with low reliance on $X'$. However, in this case, we modify an existing informative feature instead of creating a new feature.

\subsection{Subsampling Based on Unobserved Feature (\texttt{BiasSampUnobs})}

\label{subsec:unobs}
We create an augmented version of the dataset by subsampling based on a binary feature $X'$ to create confounding. We also remove this feature from modelling to reflect realistic scenarios of induced sampling bias due to unknown factors. We configure the desired data parameters ${\mu_1^e = P(Y = 1 | X' = 1)}$ and ${\mu_0^e = P(Y = 1 | X' = 0)}$. We then randomly subsample each environment for each value of $X'$ separately to achieve the desired label distribution. The algorithm for subsampling is shown in Algorithm~1. The causal graph for this augmentation is shown in Figure~\ref{fig:flow}. A practical scenario reflecting this setting is one where the degree of sampling bias differs across environments.

\begin{algorithm}
\label{alg:subsample}
\DontPrintSemicolon
 \KwData{$(x', y)$: gender and label of sample }
  \KwData{$\mu$: desired prevalence of $x'$ in environment}
  \KwData{$\tau$: current prevalence of $x'$ in environment}
  
  \KwResult{probability that the sample will be dropped}
  \uIf{$y == 1$ and $\tau > \mu$ }{ 
   \Return $1 - \frac{1 - \tau}{\tau} \cdot \frac{\mu}{1 - \mu}$ 
  }\ElseIf{$y == 0$ and $\tau < \mu$ }{ 
    \Return $1 - \frac{\tau}{1 - \tau} \cdot \frac{1 - \mu}{\mu}$
  } 
  \Return 0
  
\caption{Compute subsampling probability}
\end{algorithm}


Here, the distribution $p(Y | X')$ is not invariant across environments. If the difference between $\mu_1^e$ and $\mu_0^e$ is large, $X'$ becomes highly informative, and an ERM model would tend towards a predictor that outputs the most likely class for each value of $X'$, i.e. a classifier that outputs $\hat{Y} = \argmax_y p_{train}(Y = y | X' = f(\mathbf{X}))$, where $f$ is a model that predicts $X'$ given features $\mathbf{X}$. Because the distribution $p_{test}(Y | X')$ is vastly different from $p_{train}(Y | X')$, this confounding-reliant predictor would then have poor OOD performance. If $X'$ is set to be a protected attribute (for example, gender), depending on the settings of $\mu_1^e$ and $\mu_0^e$, this classifier could also have large performance disparities between groups. Here, the protected group would be an example of a hidden stratification~\cite{oakden2020hidden}.

Several prior connections as elicited in the related work have been made between domain generalization and potential improved statistical parity between protected attributes. Therefore, in addition to overall model performance, we also evaluate the following metrics related to algorithmic fairness. As these metrics require a binarized prediction, we choose the threshold that results in the maximum F1 score for each model.
\begin{itemize}
    \item Gap in the True Positive Rate (TPR) between the two protected groups. This corresponds to equality of opportunity for the positive class~\cite{hardt2016equality}.
    
     \item Gap in the True Negative Rate (TNR) between the two protected groups. This corresponds to equality of opportunity for the negative class~\cite{hardt2016equality}.
    
    
    \item The correlation, evaluated using the Matthews correlation coefficient~\cite{yule1912methods}, between the predicted label and the binary confounder. This roughly measures how close the learnt classifier is to a protected attribute predictor, and, in turn, how robust it is to the distribution shift.
\end{itemize}

\subsection{Subsampling Based on Observed Feature (\texttt{BiasSampObs})}
We have the identical setup as in Section \ref{subsec:unobs}. However, we now include the confounded feature in our model. This allows us to investigate the model behaviour when it has direct access to the domain-shifted feature.  

\begin{table*}[]
\caption{\label{tab:dataset_stats}Statistics of each region for the eICU in-hospital mortality prediction task and the Chest X-ray classification tasks. Label distribution for the CXR datasets are shown for the pneumonia prediction task. Detailed dataset statistics can be found in Appendix \ref{sec_app:dataset}. }
\begin{tabular}{l|lllll|llll}
                        & \multicolumn{5}{l|}{\textbf{In-Hospital Mortality (eICU)}}                                & \multicolumn{4}{l}{\textbf{Chest X-Rays (CXR)}}                                  \\ \hline
\textbf{Environment}    & \textbf{Midwest} & \textbf{West} & \textbf{Northeast} & \textbf{Missing} & \textbf{South} & \textbf{MIMIC-CXR} & \textbf{CheXpert} & \textbf{Chest-Xray8} & \textbf{PadChest} \\ \hline
\textbf{Assigned Split} & Train            & Train         & Train              & Validation       & Test           & Train              & Train             & Validation           & Test              \\
\textbf{\# Samples}     & 10,985           & 4,527         & 2,495              & 1,846            & 10,827         & 249,995            & 191,229           & 112,120              & 99,934            \\
\textbf{\% Positive}    & 9.43\%           & 14.42\%       & 13.19\%            & 12.68\%          & 11.74\%        & 7.37\%             & 2.45\%            & 1.28\%               & 4.90\%           
\end{tabular}
\end{table*}

\section{Data and Models}

We consider clinical data from two distinct data domains -- time-series data and images. We also include results for the Colored MNIST dataset in Appendix \ref{sec_app:cmnist}.

\subsection{In-Hospital Mortality (eICU)}
\label{subsec:eicu}
\paragraph{Dataset.}
The eICU collaborative research database V2.0~\cite{pollard2018eicu} consists of intensive care unit (ICU) records for over $200,000$ admissions to over $200$ hospitals across the United States. We use the cohort creation procedure for the in-hospital mortality prediction task outlined by \citet{sheikhalishahi2020benchmarking}. The goal is to predict whether a patient will die in hospital, given data from the first $48$ hours of their hospital stay. Patients who die within the first $48$ hours are removed from the cohort, as are patients who are younger than 18 or older than $89$ years of age. Patients who have more than one ICU stay only have their first stay selected. Time-series observations (labs and vitals) are grouped into 1-hour windows, with missing values imputed from the previous observation.

For each patient, we have $10$ continuous and $4$ categorical time series features, and $3$ continuous and $2$ categorical static features. A complete list of these features can be found in Table \ref{tab:eicu_features}.
The resulting dataset consists of 30,680 patients, 11.48\% of which have a positive label. Each patient is associated with a hospital, which is located in one of four regions in the United States. A small number of hospitals do not have an associated region in the database. A summary of the statistics for each region is shown in Table \ref{tab:dataset_stats}.

\paragraph{Domains.}
We use \texttt{Midwest}, \texttt{South}, and \texttt{West} as training environments, and we use \texttt{Missing} as the validation environment. We choose \texttt{South} as the test environment, as its demographics appear to be the most distinct of the five, as seen in Table \ref{tab:eicu_full_stats}. 

\paragraph{Models.}
We use a gated recurrent neural network~\cite{chung2014empirical}, with a linear classifier over the final hidden state. Categorical variables are embedded before being input to the network, and continuous features are scaled to zero mean and unit variance. Static features are appended to time-series features at each timestep.

\paragraph{Hyperparameter Search.} We use ten iterations of random search~\cite{bergstra2012random} to tune the learning rate and hyperparameters specific to each algorithm, randomizing the data splits and model initialization each time. Following the practice established by \citet{gulrajaniSearchLostDomain2020}, we select the best model of the ten with the associated model selection strategy (using AUROC as the metric), and then repeat this entire procedure five times, reporting the mean and standard deviation of metrics for the best models across the trials. This allows us to account for the variance in the hyperparameter search procedure.

\paragraph{Experiments.}
We benchmark this dataset using all of the experimental settings defined in Section \ref{sec:framework}. For \texttt{CorrNoise}, we choose $\tilde{X}$ to be the admission weight (a static continuous feature). For \texttt{BiasSampUnobs} and \texttt{BiasSampObs}, we use gender as the confounding variable. We set $\mu_M^e$ and $\mu_F^e$ to the values shown in Table \ref{tab:subsample_dg_params}.

\subsection{Chest X-rays (CXR)}
\paragraph{Dataset.}
We use four public chest X-ray (CXR) datasets: MIMIC-CXR~\cite{johnson_mimic-cxr:_2019}, CheXpert~\cite{irvin_chexpert:_2019}, Chest-Xray8~\cite{wang_chestx-ray8:_2017}, and PadChest~\cite{bustos_padchest:_2019}. Statistics for each dataset can be found in Table \ref{tab:dataset_stats}, and detailed statistics can be found in Table \ref{tab:cxr_full_stats}. Each sample consists of a chest X-ray image along with zero or more diagnostic labels.

We preprocess the data to obtain eight common labels shared between all datasets. Though some datasets contain both frontal and lateral CXR images, we use only frontal images (both PA and AP views) for our experiments to prevent presence of additional confounding in our analysis. 

\paragraph{Domains.}
We designate each dataset as its own environment. We use the PadChest dataset as the test environment because it is the only dataset from a hospital located outside of the United States, and because prior work has shown it to be the domain with the worst performance as the transfer target~\cite{poochCanWeTrust2020}.

\paragraph{Models.}
We use a DenseNet-121~\cite{huang2017densely} network, initializing with pre-trained weights from ImageNet~\cite{imagenet_cvpr09}, which has been shown to perform well on CXR classification~\cite{raghu_transfusion:_2019, bressem2020comparing}. We replace the final layer with a linear layer of the appropriate size. For training the network, all images are scaled to $224 \times 224$ and normalized to the ImageNet mean and standard deviation. We apply multiple image augmentations to the training set: flipping of the images along the horizontal axis, rotation of up to $10$ degrees, and a crop of a random size $(75\%-100\%)$ and a random aspect ratio ($3/4$ to $4/3$). 

\paragraph{Hyperparameter Search.} We use the same hyperparameter search strategy as described in Section \ref{subsec:eicu}.

\paragraph{Experiments.} 
We define two predictive setups. In the \textit{multitask} setup, we learn a network that jointly predicts the eight labels simultaneously, trained to minimize the mean of the binary cross-entropy over all tasks. For model selection, we use the average AUROC across all eight labels as the metric. In the \textit{binary} setup, we select only the pneumonia label, and learn a binary classifier to predict whether an image contains a lung infected with pneumonia. For model selection, we use AUROC as the metric. 

We benchmark this dataset for the \texttt{Base} setting using both the \textit{multitask} and \textit{binary} setups, and for the \texttt{BiasSampUnobs} and \texttt{BiasSampObs} settings using the \textit{binary} setup. We omit \texttt{CorrLabel} and \texttt{CorrNoise} here, as these shifts are not clinically meaningful for x-ray images. For the biased subsampling shifts, we use gender as the confounding variable. We set $\mu_M^e$ and $\mu_F^e$ to the values shown in Table \ref{tab:subsample_dg_params}.

\begin{table}[]
\caption{\label{tab:subsample_dg_params} Data parameters for the subsampling experiments and the resulting gender distribution.}
\begin{tabular}{l|l|r|r|r|r}
\textbf{Dataset} & \textbf{Environment} & $\bm{\mu_M}$ & $\bm{\mu_F}$ & \textbf{\% Male} & \textbf{\% Female} \\ \hline
eICU    & Midwest     & 0.8   & 0.05  & 35.7\%  & 64.3\%    \\
        & West        & 0.7   & 0.1   & 57.6\%  & 42.4\%    \\
        & Northeast   & 0.6   & 0.15  & 51.2\%  & 48.8\%    \\
        & Missing     & 0.3   & 0.3   & 50.3\%  & 49.7\%    \\
        & South       & 0.1   & 0.5   & 82.8\%  & 17.2\%    \\ \hline
CXR     & MIMIC-CXR       & 0.2   & 0.02  & 30.2\%  & 69.8\%    \\
        &  CheXpert        & 0.1   & 0.03  & 28.6\%  & 71.4\%    \\
        & Chest-Xray8         & 0.07  & 0.04  & 30.8\%  & 69.2\%    \\
        & PadChest         & 0.05  & 0.05  & 54.6\%  & 45.4\%   
\end{tabular}
\end{table}

\section{Results}

\begin{table*}[]
\caption{\label{tab:no_confounding} Performance results for \texttt{Base}. We evaluate the AUROC performances on the test environment for the base datasets. We find that no domain generalization method significantly improves model performance on CXR classification consistently. }
\resizebox{.99\textwidth}{!}{
\begin{tabular}{l|l|rr|rrrrrrrr}
\textbf{Model Selection}                                                               & \textbf{Dataset} & \textbf{OracleID} & \textbf{OracleMerged} & \textbf{ERM}         & \textbf{GroupDRO}    & \textbf{IRM} & \textbf{VREx} & \textbf{RVP} & \textbf{IGA} & \textbf{CORAL}       & \textbf{MLDG} \\ \hline
\multirow{3}{*}{\textbf{\begin{tabular}[c]{@{}l@{}}Training \\ Domains\end{tabular}}}  & eICU             & 0.852±0.009       & 0.878±0.010           & \textbf{0.879±0.015} & 0.856±0.012          & 0.870±0.011  & 0.873±0.018   & 0.866±0.014  & 0.873±0.007  & 0.870±0.020          & 0.876±0.013   \\
                                                                                       & CXR (multitask)  & 0.894±0.006       & 0.900±0.004           & \textbf{0.862±0.007} & 0.856±0.004          & 0.857±0.019  & 0.846±0.006   & 0.847±0.019  & 0.794±0.028  & 0.855±0.010          & 0.803±0.009   \\
                                                                                       & CXR (binary)     & 0.841±0.022       & 0.812±0.020           & 0.718±0.053          & 0.725±0.048          & 0.711±0.076  & 0.719±0.029   & 0.727±0.026  & 0.639±0.051  & \textbf{0.756±0.007} & 0.600±0.035   \\ \hline
\multirow{3}{*}{\textbf{\begin{tabular}[c]{@{}l@{}}Validation \\ Domain\end{tabular}}} & eICU             & 0.852±0.009       & 0.878±0.010           & \textbf{0.871±0.020} & 0.860±0.014          & 0.851±0.018  & 0.865±0.008   & 0.858±0.021  & 0.862±0.015  & 0.857±0.024          & 0.869±0.018   \\
                                                                                       & CXR (multitask)  & 0.894±0.006       & 0.900±0.004           & 0.845±0.014          & \textbf{0.860±0.009} & 0.846±0.011  & 0.844±0.012   & 0.838±0.015  & 0.780±0.015  & 0.850±0.009          & 0.805±0.023   \\
                                                                                       & CXR (binary)     & 0.841±0.022       & 0.812±0.020           & 0.697±0.046          & 0.730±0.034          & 0.693±0.029  & 0.712±0.034   & 0.725±0.051  & 0.643±0.055  & \textbf{0.735±0.017} & 0.578±0.039  
\end{tabular}
}
\end{table*}

\begin{table*}[]
\caption{\label{tab:corr_gaussian}  Performance results for \texttt{CorrLabel} and \texttt{CorrNoise}. We evaluate the AUROC performances on the South environment in eICU mortality prediction with addition of a corrupted version of the label as a feature (\texttt{CorrLabel}) and addition of correlated Gaussian noise (\texttt{CorrNoise}). We find that model performance improves as the the distance between training environments increases, and that there exist significant performance gains for domain generalization methods in cases where the spurious correlation is extreme.
}
\resizebox{.99\textwidth}{!}{
\begin{tabular}{l|l|rr|r|rrrrrrrr}
\textbf{Model Selection}                                                              & \textbf{Setting}                            & \textbf{OracleID}            & \textbf{OracleMerged} & \textbf{ERM Unaug}           & \textbf{ERM} & \textbf{GroupDRO}    & \textbf{IRM}         & \textbf{VREx} & \textbf{RVP}         & \textbf{IGA}         & \textbf{CORAL}       & \textbf{MLDG}        \\ \hline
\multirow{7}{*}{\textbf{\begin{tabular}[c]{@{}l@{}}Training\\ Domains\end{tabular}}}  & CorrLabel ($\beta = 0.1$)                     & \multirow{3}{*}{0.963±0.006} & 0.767±0.021           & \multirow{7}{*}{0.879±0.015} & 0.305±0.048  & 0.317±0.022          & 0.347±0.041          & 0.349±0.067   & 0.399±0.026          & \textbf{0.400±0.057} & 0.359±0.030          & 0.350±0.051          \\
                                                                                      & CorrLabel ($\beta = 0.3$)                     &                              & 0.862±0.011           &                              & 0.694±0.034  & 0.660±0.046          & 0.704±0.031          & 0.702±0.025   & 0.688±0.028          & \textbf{0.726±0.028} & 0.709±0.037          & 0.687±0.025          \\
                                                                                      & CorrLabel ($\beta = 0.5$)                     &                              & 0.911±0.005           &                              & 0.865±0.012  & 0.845±0.015          & 0.862±0.013 & 0.871±0.018   & 0.862±0.008          & 0.869±0.008          & \textbf{0.872±0.010} & 0.857±0.014          \\ \cline{2-4} \cline{6-13} 
                                                                                      & CorrNoise ($\beta = 1.0, \delta = 0.5$) & \multirow{4}{*}{0.959±0.008} & 0.794±0.023           &                              & 0.388±0.047  & 0.422±0.046          & 0.386±0.021          & 0.374±0.031   & 0.418±0.037          & 0.404±0.042 & \textbf{0.440±0.050} & 0.410±0.020          \\
                                                                                      & CorrNoise ($\beta = 1.0, \delta = 1.0$) &                              & 0.826±0.026           &                              & 0.556±0.032  & 0.614±0.080          & 0.557±0.041          & 0.571±0.013   & \textbf{0.655±0.030} & 0.565±0.064          & 0.600±0.033          & 0.548±0.014          \\
                                                                                      & CorrNoise ($\beta = 2.0, \delta = 0.5$) &                              & 0.717±0.022           &                              & 0.209±0.027  & 0.214±0.048          & 0.207±0.020          & 0.193±0.023   & 0.191±0.018          & 0.200±0.026 & \textbf{0.234±0.027} & 0.199±0.028          \\
                                                                                      & CorrNoise ($\beta = 2.0, \delta = 1.0$) &                              & 0.730±0.024           &                              & 0.244±0.025  & 0.253±0.031          & 0.254±0.028          & 0.245±0.025   & 0.251±0.019          & 0.263±0.027 & 0.279±0.033          & \textbf{0.281±0.027} \\ \hline
\multirow{7}{*}{\textbf{\begin{tabular}[c]{@{}l@{}}Validation\\ Domain\end{tabular}}} & CorrLabel ($\beta = 0.1$)                     & \multirow{3}{*}{0.963±0.006} & 0.767±0.021           & \multirow{7}{*}{0.871±0.020} & 0.678±0.087  & 0.677±0.065 & \textbf{0.733±0.016} & 0.612±0.142   & 0.715±0.045          & 0.683±0.069          & 0.689±0.082          & 0.690±0.056          \\
                                                                                      & CorrLabel ($\beta = 0.3$)                     &                              & 0.862±0.011           &                              & 0.697±0.045  & 0.724±0.024          & 0.748±0.032          & 0.716±0.025   & 0.684±0.052          & \textbf{0.757±0.035} & 0.699±0.021          & 0.717±0.015          \\
                                                                                      & CorrLabel ($\beta = 0.5$)                     &                              & 0.911±0.005           &                              & 0.865±0.013  & 0.862±0.014          & \textbf{0.868±0.016} & 0.845±0.023   & 0.855±0.017          & 0.865±0.010          & 0.862±0.007          & 0.860±0.009          \\ \cline{2-4} \cline{6-13} 
                                                                                      & CorrNoise ($\beta = 1.0, \delta = 0.5$) & \multirow{4}{*}{0.959±0.008} & 0.794±0.023           &                              & 0.446±0.139  & 0.415±0.057          & 0.463±0.097          & 0.494±0.085   & 0.466±0.049          & \textbf{0.596±0.132} & 0.497±0.101          & 0.414±0.071          \\
                                                                                      & CorrNoise ($\beta = 1.0, \delta = 1.0$) &                              & 0.826±0.026           &                              & 0.561±0.090  & 0.585±0.057          & 0.522±0.020          & 0.566±0.099   & \textbf{0.669±0.062} & 0.641±0.084          & 0.591±0.058          & 0.541±0.053          \\
                                                                                      & CorrNoise ($\beta = 2.0, \delta = 0.5$) &                              & 0.717±0.022           &                              & 0.492±0.181  & 0.489±0.110          & 0.423±0.181          & 0.395±0.075   & 0.385±0.060          & \textbf{0.513±0.087} & 0.503±0.053          & 0.467±0.112          \\
                                                                                      & CorrNoise ($\beta = 2.0, \delta = 1.0$) &                              & 0.730±0.024           &                              & 0.506±0.149  & 0.417±0.169          & 0.436±0.171          & 0.405±0.110   & 0.414±0.128          & \textbf{0.546±0.152} & 0.366±0.111          & 0.347±0.087         
\end{tabular}
}
\end{table*}

\subsection{Performance on Base Datasets}
\paragraph{\textbf{ERM Peforms Well Across Targets and Shifts.}}
Table \ref{tab:no_confounding} shows the performance of each of the domain generalization methods on the test environment. First, comparing the performance of the OracleID and ERM methods, we note that for the CXR setups, there is indeed a statistically significant drop in performance when a model is trained on PadChest, versus when a model is transferred to PadChest. 

Surprisingly, the performance of ERM on the eICU test set is actually on-par with the oracles, indicating that the \texttt{South} environment is likely not OOD. Therefore, it is not fair to make conclusions about the performance of domain generalization methods based on their performance on \texttt{Base} eICU.

In the CXR setting, none of the domain generalization methods consistently outperform ERM, though CORAL performs quite well in the binary task, and many of the methods perform significantly worse than ERM. This result is consistent with prior work~\cite{kohWILDSBenchmarkIntheWild2020a, gulrajaniSearchLostDomain2020}. 

\begin{figure*}
    \centering
    \includegraphics[width=0.8\textwidth]{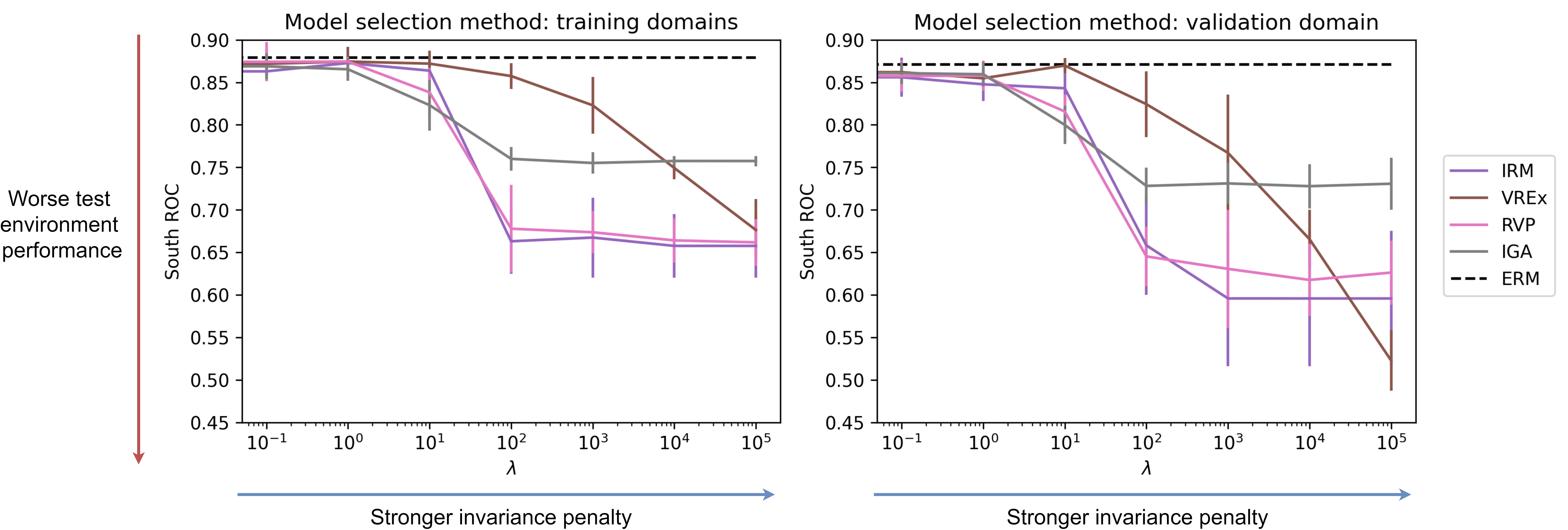}
    \caption{For the \texttt{Base} eICU mortality prediction dataset, we select methods that balance the ERM loss with some invariance loss term using some hyperparameter $\lambda$. We vary $\lambda$ from a small value (where the loss function is equivalent to ERM) to a large value (where the training environment invariances are strongly enforced). We find that defaulting to ERM yields the best test environment performance. }
    \label{fig:vary_lambda}
\end{figure*}

\paragraph{\textbf{Enforcing Invariance Can Harm Performance.}}
We examine the methods that have a tunable $\lambda$ parameter that balances the standard ERM loss with some invariance enforcing loss. This evaluation helps to investigate whether domain generalization methods fall-back to ERM (i.e. small $\lambda$), and is lacking in prior benchmarks. As shown in Figure \ref{fig:vary_lambda}, We vary $\lambda$, and find that, in the case where the test environment is not OOD, enforcing invariances in the model can actually significantly hurt test domain performance. 

\subsection{Performance Under Synthetic Domain Shift}
We examine the results of \texttt{CorrLabel} and \texttt{CorrNoise} (Table \ref{tab:corr_gaussian}), and \texttt{BiasSampUnobs} and \texttt{BiasSampObs}  (Table \ref{tab:roc_subsample}).

\paragraph{\textbf{Domain Generalization Shows Limited Effectiveness Under Extreme Spurious Correlations.} }
There are indeed scenarios where domain generalization methods outperform ERM, but improvements are limited, and only become significant when the strength of the spurious correlation is extreme. In such cases, ERM is completely reliant on the spurious correlation, and performs worse than chance on the test environment where the spurious correlation is flipped. This provides the opportunity for a performance gain for the domain generalization methods. However, even in such cases, the performance of domain generalization methods is still quite poor relative to the unaugmented case -- which represents the performance of an ideal ERM model that ignores the spurious correlation. We note that although the OracleID model has exceptionally high performance in the experiments, it is completely reliant on the spurious correlation in the test environment, and would thus transfer very poorly.

\paragraph{\textbf{Validation Environment Model Selection is More Robust.} }
Next, we observe that, in almost all cases, model selection on the validation environment yields better performance than using the training domains. Since we specifically designed the validation environment to have an intermediate level of spuriousness between the training and test domains, this result is to be expected.

\paragraph{\textbf{Increased Training Diversity Improves Generalization.} }
Finally, for the correlated noise experiment in Table \ref{tab:corr_gaussian}, we observe that increasing the diversity between the environments by increasing $\delta$ significantly increases performance for the large majority of models. When the gap between the environments increase, it is easier for the models to detect the spurious correlation, as relying on the spurious correlation would lead to comparably worse training loss, resulting in better generalization.

\subsection{Domain Generalization and Fairness Under Sampling Bias}
\begin{table*}[!h]
\caption{\label{tab:roc_subsample} Performance results for \texttt{BiasSampUnobs} and \texttt{BiasSampObs}. We evaluate the test environment AUROC in subsampling experiments with eICU and CXR datasets. We notice that observing the subsampled feature reduces generalization performance, and that domain generalization methods do not consistently significantly outperform ERM.}
\resizebox{.99\textwidth}{!} {
\begin{tabular}{l|l|l|rr|rrrrrrrr}
\textbf{Dataset}                                                                 & \textbf{\begin{tabular}[c]{@{}l@{}}Selection\\ Method\end{tabular}} & \textbf{Observed} & \textbf{OracleID} & \textbf{OracleMerged} & \textbf{ERM}         & \textbf{GroupDRO}    & \textbf{IRM} & \textbf{VREx}        & \textbf{RVP}         & \textbf{IGA}         & \textbf{CORAL}       & \textbf{MLDG} \\ \hline
\multirow{4}{*}{\textbf{eICU}}                                                   & \multirow{2}{*}{\textbf{Training}}                                  & \textbf{No}       & 0.886±0.015       & 0.849±0.020           & 0.766±0.013          & 0.776±0.020          & 0.760±0.021  & 0.757±0.034          & 0.750±0.024          & \textbf{0.782±0.026} & 0.771±0.033          & 0.756±0.020   \\
                                                                                 &                                                                     & \textbf{Yes}      & 0.896±0.008       & 0.826±0.024           & 0.652±0.010          & \textbf{0.657±0.032} & 0.648±0.044  & 0.654±0.058          & \textbf{0.657±0.033} & 0.618±0.071          & 0.654±0.030          & 0.656±0.041   \\ \cline{2-13} 
                                                                                 & \multirow{2}{*}{\textbf{Validation}}                                & \textbf{No}       & 0.886±0.015       & 0.849±0.020           & 0.778±0.034          & 0.769±0.047          & 0.769±0.010  & 0.765±0.022          & \textbf{0.787±0.010} & 0.746±0.032          & 0.778±0.032          & 0.775±0.027   \\
                                                                                 &                                                                     & \textbf{Yes}      & 0.896±0.008       & 0.826±0.024           & 0.689±0.023          & 0.692±0.018          & 0.690±0.054  & \textbf{0.718±0.033} & 0.712±0.042          & 0.608±0.102          & 0.685±0.021          & 0.672±0.029   \\ \hline
\multirow{4}{*}{\textbf{\begin{tabular}[c]{@{}l@{}}CXR\\ (Binary)\end{tabular}}} & \multirow{2}{*}{\textbf{Training}}                                  & \textbf{No}       & 0.840±0.010       & 0.811±0.015           & 0.640±0.032          & 0.648±0.049          & 0.605±0.040  & 0.640±0.029          & 0.622±0.035          & 0.571±0.013          & \textbf{0.653±0.034} & 0.590±0.085   \\
                                                                                 &                                                                     & \textbf{Yes}      & 0.844±0.006       & 0.817±0.006           & \textbf{0.669±0.043} & 0.629±0.023          & 0.639±0.027  & 0.626±0.044          & 0.619±0.045          & 0.558±0.046          & 0.631±0.028          & 0.567±0.044   \\ \cline{2-13} 
                                                                                 & \multirow{2}{*}{\textbf{Validation}}                                & \textbf{No}       & 0.840±0.010       & 0.811±0.015           & 0.624±0.041          & 0.632±0.028          & 0.630±0.013  & \textbf{0.682±0.038} & 0.640±0.029          & 0.639±0.048          & 0.637±0.040          & 0.611±0.051   \\
                                                                                 &                                                                     & \textbf{Yes}      & 0.844±0.006       & 0.817±0.006           & 0.658±0.025          & 0.655±0.023          & 0.621±0.073  & 0.656±0.033          & 0.650±0.047          & \textbf{0.669±0.038} & 0.615±0.053          & 0.640±0.044  
\end{tabular}
}
\end{table*}

\begin{figure*}
\centering
\includegraphics[width=0.8\linewidth]{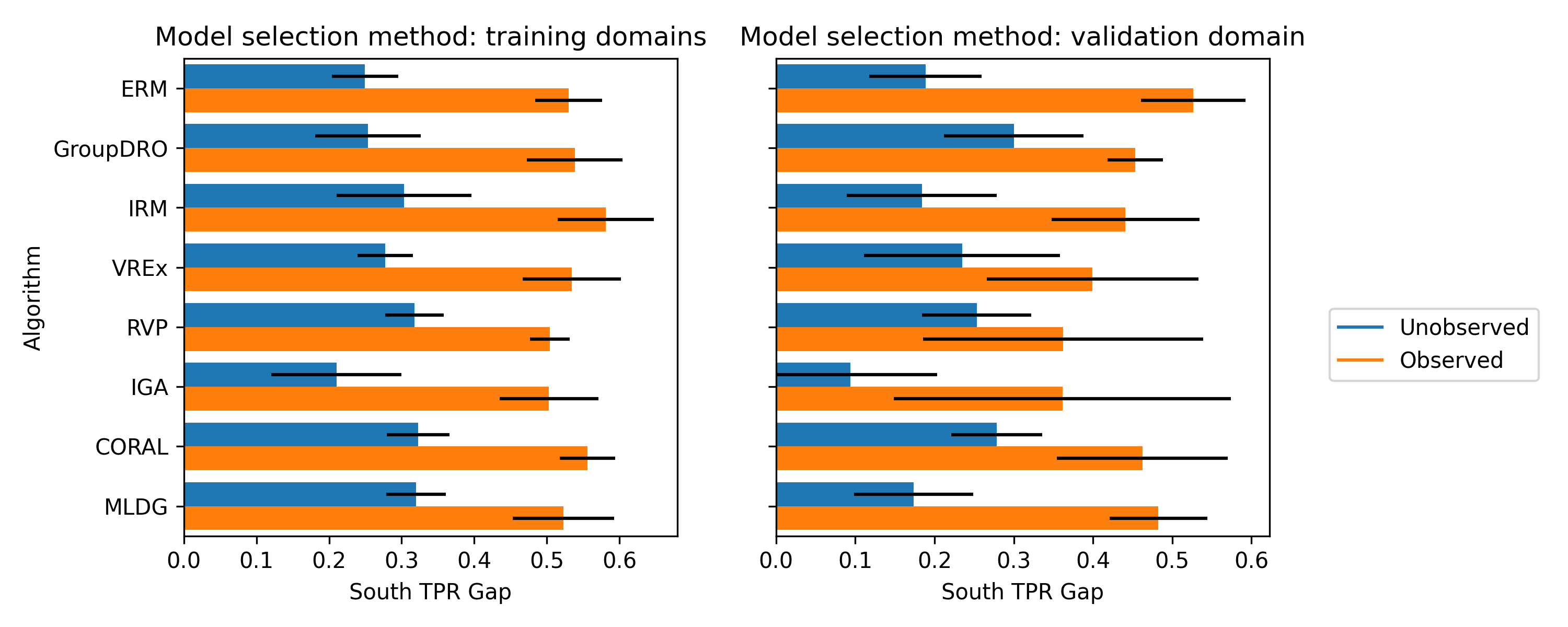}
\caption{\label{fig:tpr_subsample} TPR gaps for \texttt{BiasSampUnobs} and \texttt{BiasSampObs} on eICU. We evaluate the test environment true positive rate gaps (M-F) in subsampling experiments with the eICU dataset. We notice that observing the subsampled feature greatly increases the TPR disparity. Though there exist instances where domain generalization methods have lower disparity than ERM, the corresponding models also generally have lower AUROC. Corresponding results for CXR are shown in Table \ref{tab:tpr_subsample}.}
\end{figure*}



\paragraph{\textbf{Domain Generalization Does Not Produce Fairer Classifiers with Better Performance.}}
First, we observe from Table~\ref{tab:roc_subsample}, similar to our results on the base datasets in Table~\ref{tab:no_confounding}, that domain generalization methods do not show significant improvements in overall performance over ERM. Next, looking at the TPR gaps in Figure~\ref{fig:tpr_subsample} and Table~\ref{tab:tpr_subsample}, we find that few models have significantly lower disparity than ERM, and the models that do have much lower overall utility. There do not appear to be models that improve on both overall performance and fairness over ERM. This is also observed in the TNR gaps (Table \ref{tab:tnr_subsample}). Given that a model which ignores the spurious gender correlation should be both fairer and have better performance than ERM, we conclude that domain generalization methods are not capable of overcoming spurious correlations induced through subsampling.

\paragraph{\textbf{Observing Confounding Can Reduce Fairness.}}
Interestingly, we see that both domain generalization algorithms and ERM produce classifiers with significantly worse fairness, along with worse overall performance, when given the value of the subsampled feature. We observe in Table \ref{tab:phi_subsample} that the correlation coefficient between gender and the model prediction is also significantly higher when the protected group is given to the model. It appears that the model becomes more reliant on the spurious correlation when its value is directly provided, resulting in poor performance and fairness under distribution shift.

\section{Discussion}
\subsection{Disparity Between Real World and Manually Confounded Data}
We offer several hypothesis for why domain generalization seems to perform well in limited settings compared with ERM on the manually augmented data, but performs much worse on real-world medical imaging data, as well as various real-world benchmarks from prior work~\cite{kohWILDSBenchmarkIntheWild2020a}.

First, the spurious correlations that we introduce are fairly simple and extreme in magnitude -- there is often one single variable which the model should avoid in order to achieve a decent result. In the real world, the spurious correlations that exist are much more subtle and complex, and it is not as simple to isolate it in the causal graph as in our scenarios. Secondly, prior work has demonstrated certain synthetic settings where IRM provably   recovers a suboptimal predictor \cite{kamath2021does}. The sample complexity of IRM~\cite{ahuja2020empirical} versus ERM could also be relevant. However, in the real world where the underlying data generating distribution is unknown, the degree to which these factors contribute to our observations is unclear.  

Finally, prior theoretical work into the diversity requirement for the environments in IRM has shown that IRM will fail unless the training environments ``cover'' the space of all possible environments~\cite{rosenfeldRisksInvariantRisk2020}. In our synthetic augmentation scenarios, we can easily tune the data hyperparameters to increase the space covered by the environments. However, in the real world, where the number of spurious and invariant features are unknown, it is unclear what diversity requirement is needed, or how many environments would be required. Nonetheless such transparent evaluation with added confounding is critical to expose these limitations.


\subsection{Domain Generalization and Fairness Under Sampling Bias} 

There are some findings of note in our analysis of fairness in experiments with subsampling. We observe that including the protected group as a feature in the classifier leads to worse performance, more unfair predictions, and greater correlation between model prediction and the gender attribute for both ERM and domain generalization methods. This is consistent with prior findings which show that the inclusion of spurious correlations can have significant effects on accuracy and group fairness \cite{khani2020removing}. 

Our results demonstrate that domain generalization methods do not provide improved performance along with improved fairness guarantees over ERM in sampling bias experiments, both in cases with awareness of the sensitive attribute and without knowledge of the protected feature. We do note there exist models which trade-off model performance for increased fairness. This trade-off has also been observed in the supervised learning setting \cite{menon2018cost}. However, as a random binary classifier is perfectly fair, the real-world utility of these models should be determined on a case-by-case basis.  

Our results appear inconsistent with prior work in that ~\citet{creager2020exchanging} prove a direct relationship between group sufficiency and IRM objective. However, ~\citet{creager2020exchanging} demonstrate this theoretical result in the setting where the sensitive attribute is taken to be the environment label.  ~\citet{adragna2020fairness} show empirically that IRM can overcome the fairness impairment faced by ERM when a spurious correlation is introduced between the label and certain demographic groups through label flipping. Both differ from our experimental setup which studies the fairness of domain generalization and ERM in the context of sampling bias -- where groups have varying label distributions across different environments. 

In this work, we study fairness provided by domain generalization methods on healthcare datasets according to common fairness metrics in machine learning such as equalized odds, we emphasize that such fairness criteria may not be relevant nor particularly useful in healthcare datasets where class often denotes diagnosis. For this reason, we suggest for future work that domain generalization in the clinical sector be evaluated according to other ethical and fairness criteria more suited to healthcare. 

\subsection{Best Practices for Domain Generalization in Medicine} 

From our evaluation of domain generalization using limited but publicly available healthcare datasets, we provide the following broad insights for applying domain generalization in medicine. First, very few existing benchmarks compare benefits of domain generalization methods to the oracle baseline, where test set data is observed. Though this baseline is impractical from a domain generalization perspective, in reality, a hospital could easily choose to train and deploy a model only on their data, instead of transferring from publicly available datasets. We find that, though this model would learn spurious correlations that exist within that hospital, this approach outperforms domain generalization in almost all cases.


It is also important to consider the test environments for which the model will be deployed. In the domain generalization setup, there is no prior knowledge about how the test environments will look like during training time. However, this is not always the case in the real world. If there is a guarantee that the model trained will only be deployed at large hospitals only in the US, and temporal domain shift is not a factor, a simple ERM model could perform quite well, while relying on spurious correlations consistent across the US environments. In fact, in such cases, the test environment might not even be OOD, as in the base eICU example. This is more likely to be true when the majority of observed features (such as vitals or lab tests) tend to be invariant across demographics. In this scenario, careless application of domain generalization methods (with improper hyperparmeter tuning) could actually lead to worse model performance compared to ERM.

If, instead, the model created has the potential to be deployed in all regions throughout the world, it is then critical to learn a model that does not rely on US-specific spurious correlations. Domain generalization could potentially be useful in this case, where an invariant model is learnt in exchange for worse performance at sites in the US. In such cases, it is important to train on a set of environments that is as diverse as possible. This further suggests that without real diversity in training environments, learning models that are truly invariant to such spurious correlations is not possible with existing methods. It is also highly beneficial to conduct model selection using an environment, or a combination of environments, closest to where the model will be deployed.

In the field of medicine specifically, there already exist many known causal effects between various observed features~\cite{nordon2019building, etminan2020using}. Working with domain experts to delve into existing causal relationships in tabular data can provide invaluable insight both for constructing and benchmarking invariant models.

Finally, when considering the performance of a domain generalization method, it is important to look past its performance on Colored MNIST, as state-of-the-art performance in these datasets appears to have little correlation with performance on real-world data, likely due to their model selection using the test environment (see Appendix \ref{sec_app:cmnist}). Instead, it is important to consider their performance on a large variety of realistic benchmarks such as DomainBed~\cite{gulrajaniSearchLostDomain2020}, WILDS~\cite{kohWILDSBenchmarkIntheWild2020a}, or our clinical framework.


\section{Conclusion}
Clinical models trained on one hospital or region typically degrade in performance in the presence of domain shift~\cite{zechVariableGeneralizationPerformance2018, poochCanWeTrust2020, cohenLimitsCrossdomainGeneralization2020,martenssonReliabilityDeepLearning2020, albadawyDeepLearningSegmentation2018,stackeCloserLookDomain2019, stackeMeasuringDomainShift2020, thagaardCanYouTrust2020}. In this paper, we evaluated the performance of eight domain generalization methods on their ability to generalize to an unseen test environment for typical clinical datasets. We find, consistent with prior work on general image datasets~\cite{gulrajaniSearchLostDomain2020}, that these methods do not consistently exhibit significantly improved performance on chest X-ray datasets over empirical risk minimization. We then propose a framework for manually introducing realistic spurious correlations to the dataset, and find that there exist cases where domain generalization significantly outperforms empirical risk minimization. We observe no consistent improvement in fairness along with performance in the presence of sampling bias.

We believe that the results we have shown motivates the need for further testing of the failure and success modes of domain generalization in clinical settings, as well as theoretical justifications for the disparity between their performance on artificial shifts versus real-world shifts. We reiterate the message by \citet{gulrajaniSearchLostDomain2020} that the model selection strategy is an integral part of a domain generalization method, and echo the sentiment by \citet{kohWILDSBenchmarkIntheWild2020a} for more realistic benchmarks for evaluating real-world domain shifts. We believe that our empirical framework that introduces synthetic domain shifts and sampling bias will prove to be useful starting step for stress-testing novel domain generalization methods, as well as inspire further work in domain generalization in medicine.

\section*{Acknowledgements}
We would like to thank Taylor Killian and Nathan Ng for their feedback. Dr. Marzyeh Ghassemi is funded in part by Microsoft Research, a Canadian CIFAR AI Chair held at the Vector Institute, a Tier 2 Canada Research Council Chair, and an NSERC Discovery Grant. We also acknowledge NSERC (funding number PDF-516984). Resources used in preparing this research were provided, in part, by the Province of Ontario, the Government of Canada through CIFAR, and
companies sponsoring the Vector Institute.

\bibliographystyle{plainnat}
\bibliography{references}

\begin{thebibliography}{86}
\providecommand{\natexlab}[1]{#1}
\providecommand{\url}[1]{\texttt{#1}}
\expandafter\ifx\csname urlstyle\endcsname\relax
  \providecommand{\doi}[1]{doi: #1}\else
  \providecommand{\doi}{doi: \begingroup \urlstyle{rm}\Url}\fi

\bibitem[Adragna et~al.(2020)Adragna, Creager, Madras, and
  Zemel]{adragna2020fairness}
Robert Adragna, Elliot Creager, David Madras, and Richard Zemel.
\newblock Fairness and robustness in invariant learning: A case study in
  toxicity classification, 2020.

\bibitem[Ahuja et~al.(2020{\natexlab{a}})Ahuja, Shanmugam, Varshney, and
  Dhurandhar]{ahuja2020invariant}
Kartik Ahuja, Karthikeyan Shanmugam, Kush Varshney, and Amit Dhurandhar.
\newblock Invariant risk minimization games.
\newblock \emph{arXiv preprint arXiv:2002.04692}, 2020{\natexlab{a}}.

\bibitem[Ahuja et~al.(2020{\natexlab{b}})Ahuja, Wang, Dhurandhar, Shanmugam,
  and Varshney]{ahuja2020empirical}
Kartik Ahuja, Jun Wang, Amit Dhurandhar, Karthikeyan Shanmugam, and Kush~R
  Varshney.
\newblock Empirical or invariant risk minimization? a sample complexity
  perspective.
\newblock \emph{arXiv preprint arXiv:2010.16412}, 2020{\natexlab{b}}.

\bibitem[Akbarian et~al.(2020)Akbarian, Seyyed-Kalantari, Khalvati, and
  Dolatabadi]{akbarian2020evaluating}
Sina Akbarian, Laleh Seyyed-Kalantari, Farzad Khalvati, and Elham Dolatabadi.
\newblock Evaluating knowledge transfer in neural network for medical images.
\newblock \emph{arXiv preprint arXiv:2008.13574}, 2020.

\bibitem[AlBadawy et~al.(2018)AlBadawy, Saha, and
  Mazurowski]{albadawyDeepLearningSegmentation2018}
Ehab~A. AlBadawy, Ashirbani Saha, and Maciej~A. Mazurowski.
\newblock Deep learning for segmentation of brain tumors: {{Impact}} of
  cross-institutional training and testing.
\newblock \emph{Medical Physics}, 45\penalty0 (3):\penalty0 1150--1158, March
  2018.
\newblock ISSN 2473-4209.
\newblock \doi{10.1002/mp.12752}.

\bibitem[Arjovsky et~al.(2019)Arjovsky, Bottou, Gulrajani, and
  {Lopez-Paz}]{arjovskyInvariantRiskMinimization2019}
Martin Arjovsky, L{\'e}on Bottou, Ishaan Gulrajani, and David {Lopez-Paz}.
\newblock Invariant {{Risk Minimization}}.
\newblock \emph{arXiv:1907.02893 [cs, stat]}, July 2019.

\bibitem[Bandi et~al.(2018)Bandi, Geessink, Manson, Van~Dijk, Balkenhol,
  Hermsen, Bejnordi, Lee, Paeng, Zhong, et~al.]{bandi2018detection}
Peter Bandi, Oscar Geessink, Quirine Manson, Marcory Van~Dijk, Maschenka
  Balkenhol, Meyke Hermsen, Babak~Ehteshami Bejnordi, Byungjae Lee, Kyunghyun
  Paeng, Aoxiao Zhong, et~al.
\newblock From detection of individual metastases to classification of lymph
  node status at the patient level: the camelyon17 challenge.
\newblock \emph{IEEE transactions on medical imaging}, 38\penalty0
  (2):\penalty0 550--560, 2018.

\bibitem[Beam and Kohane(2018)]{beam2018big}
Andrew~L Beam and Isaac~S Kohane.
\newblock Big data and machine learning in health care.
\newblock \emph{Jama}, 319\penalty0 (13):\penalty0 1317--1318, 2018.

\bibitem[Bellot and van~der Schaar(2020)]{bellot2020generalization}
Alexis Bellot and Mihaela van~der Schaar.
\newblock Generalization and invariances in the presence of unobserved
  confounding.
\newblock \emph{arXiv preprint arXiv:2007.10653}, 2020.

\bibitem[Ben-Tal et~al.(2009)Ben-Tal, El~Ghaoui, and Nemirovski]{ben2009robust}
Aharon Ben-Tal, Laurent El~Ghaoui, and Arkadi Nemirovski.
\newblock \emph{Robust optimization}.
\newblock Princeton university press, 2009.

\bibitem[Benton et~al.(2020)Benton, Finzi, Izmailov, and
  Wilson]{bentonLearningInvariancesNeural2020}
Gregory Benton, Marc Finzi, Pavel Izmailov, and Andrew~Gordon Wilson.
\newblock Learning {{Invariances}} in {{Neural Networks}}.
\newblock \emph{arXiv:2010.11882 [cs, stat]}, December 2020.

\bibitem[Bergstra and Bengio(2012)]{bergstra2012random}
James Bergstra and Yoshua Bengio.
\newblock Random search for hyper-parameter optimization.
\newblock \emph{The Journal of Machine Learning Research}, 13\penalty0
  (1):\penalty0 281--305, 2012.

\bibitem[Bressem et~al.(2020)Bressem, Adams, Erxleben, Hamm, Niehues, and
  Vahldiek]{bressem2020comparing}
Keno~K Bressem, Lisa Adams, Christoph Erxleben, Bernd Hamm, Stefan Niehues, and
  Janis Vahldiek.
\newblock Comparing different deep learning architectures for classification of
  chest radiographs.
\newblock \emph{arXiv preprint arXiv:2002.08991}, 2020.

\bibitem[Bustos et~al.(2019)Bustos, Pertusa, Salinas, and de~la
  Iglesia-Vayá]{bustos_padchest:_2019}
Aurelia Bustos, Antonio Pertusa, Jose-Maria Salinas, and Maria de~la
  Iglesia-Vayá.
\newblock {PadChest}: {A} large chest x-ray image dataset with multi-label
  annotated reports.
\newblock \emph{arXiv:1901.07441 [cs, eess]}, 2019.

\bibitem[Carlucci et~al.(2019)Carlucci, D'Innocente, Bucci, Caputo, and
  Tommasi]{carlucci2019domain}
Fabio~M Carlucci, Antonio D'Innocente, Silvia Bucci, Barbara Caputo, and
  Tatiana Tommasi.
\newblock Domain generalization by solving jigsaw puzzles.
\newblock In \emph{Proceedings of the IEEE Conference on Computer Vision and
  Pattern Recognition}, pages 2229--2238, 2019.

\bibitem[Castro et~al.(2020)Castro, Walker, and
  Glocker]{castroCausalityMattersMedical2020}
Daniel~C. Castro, Ian Walker, and Ben Glocker.
\newblock Causality matters in medical imaging.
\newblock \emph{Nature Communications}, 11\penalty0 (1):\penalty0 3673, July
  2020.
\newblock ISSN 2041-1723.
\newblock \doi{10.1038/s41467-020-17478-w}.

\bibitem[Challen et~al.(2019)Challen, Denny, Pitt, Gompels, Edwards, and
  {Tsaneva-Atanasova}]{challenArtificialIntelligenceBias2019}
Robert Challen, Joshua Denny, Martin Pitt, Luke Gompels, Tom Edwards, and
  Krasimira {Tsaneva-Atanasova}.
\newblock Artificial intelligence, bias and clinical safety.
\newblock \emph{BMJ Quality \& Safety}, 28\penalty0 (3):\penalty0 231--237,
  March 2019.
\newblock ISSN 2044-5415, 2044-5423.
\newblock \doi{10.1136/bmjqs-2018-008370}.

\bibitem[Chen et~al.(2020)Chen, Pierson, Rose, Joshi, Ferryman, and
  Ghassemi]{chen2020ethical}
Irene~Y Chen, Emma Pierson, Sherri Rose, Shalmali Joshi, Kadija Ferryman, and
  Marzyeh Ghassemi.
\newblock Ethical machine learning in health.
\newblock \emph{arXiv preprint arXiv:2009.10576}, 2020.

\bibitem[Choe et~al.(2020)Choe, Ham, and Park]{choeEmpiricalStudyInvariant2020}
Yo~Joong Choe, Jiyeon Ham, and Kyubyong Park.
\newblock An {{Empirical Study}} of {{Invariant Risk Minimization}}.
\newblock \emph{arXiv:2004.05007 [cs, stat]}, July 2020.

\bibitem[Chung et~al.(2014)Chung, Gulcehre, Cho, and
  Bengio]{chung2014empirical}
Junyoung Chung, Caglar Gulcehre, KyungHyun Cho, and Yoshua Bengio.
\newblock Empirical evaluation of gated recurrent neural networks on sequence
  modeling.
\newblock \emph{arXiv preprint arXiv:1412.3555}, 2014.

\bibitem[Cohen et~al.(2020)Cohen, Hashir, Brooks, and
  Bertrand]{cohenLimitsCrossdomainGeneralization2020}
Joseph~Paul Cohen, Mohammad Hashir, Rupert Brooks, and Hadrien Bertrand.
\newblock On the limits of cross-domain generalization in automated {{X}}-ray
  prediction.
\newblock \emph{arXiv:2002.02497 [cs, eess, q-bio, stat]}, May 2020.

\bibitem[Cotter et~al.(2019)Cotter, Gupta, Jiang, Srebro, Sridharan, Wang,
  Woodworth, and You]{cotter2019fairness}
Andrew Cotter, Maya Gupta, Heinrich Jiang, Nathan Srebro, Karthik Sridharan,
  Serena Wang, Blake Woodworth, and Seungil You.
\newblock Training well-generalizing classifiers for fairness metrics and other
  data-dependent constraints.
\newblock In Kamalika Chaudhuri and Ruslan Salakhutdinov, editors,
  \emph{Proceedings of the 36th International Conference on Machine Learning},
  volume~97 of \emph{Proceedings of Machine Learning Research}, pages
  1397--1405, Long Beach, California, USA, 09--15 Jun 2019. PMLR.
\newblock URL \url{http://proceedings.mlr.press/v97/cotter19b.html}.

\bibitem[Creager et~al.(2020)Creager, Jacobsen, and
  Zemel]{creager2020exchanging}
Elliot Creager, Jörn-Henrik Jacobsen, and Richard Zemel.
\newblock Exchanging lessons between algorithmic fairness and domain
  generalization, 2020.

\bibitem[Deng et~al.(2009)Deng, Dong, Socher, Li, Li, and
  Fei-Fei]{imagenet_cvpr09}
J.~Deng, W.~Dong, R.~Socher, L.-J. Li, K.~Li, and L.~Fei-Fei.
\newblock {ImageNet: A Large-Scale Hierarchical Image Database}.
\newblock In \emph{CVPR09}, 2009.

\bibitem[Deng et~al.(2020)Deng, Ding, Dwork, Hong, Parmigiani, Patil, and
  Sur]{deng2020representation}
Zhun Deng, Frances Ding, Cynthia Dwork, Rachel Hong, Giovanni Parmigiani,
  Prasad Patil, and Pragya Sur.
\newblock Representation via representations: Domain generalization via
  adversarially learned invariant representations, 2020.

\bibitem[Dong et~al.(2018)Dong, Kampffmeyer, Liang, Wang, Dai, and
  Xing]{dong2018unsupervised}
Nanqing Dong, Michael Kampffmeyer, Xiaodan Liang, Zeya Wang, Wei Dai, and Eric
  Xing.
\newblock Unsupervised domain adaptation for automatic estimation of
  cardiothoracic ratio.
\newblock In \emph{International conference on medical image computing and
  computer-assisted intervention}, pages 544--552. Springer, 2018.

\bibitem[Dou et~al.(2019)Dou, de~Castro, Kamnitsas, and Glocker]{dou2019domain}
Qi~Dou, Daniel~Coelho de~Castro, Konstantinos Kamnitsas, and Ben Glocker.
\newblock Domain generalization via model-agnostic learning of semantic
  features.
\newblock In \emph{Advances in Neural Information Processing Systems}, pages
  6450--6461, 2019.

\bibitem[Etminan et~al.(2020)Etminan, Collins, and
  Mansournia]{etminan2020using}
Mahyar Etminan, Gary~S Collins, and Mohammad~Ali Mansournia.
\newblock Using causal diagrams to improve the design and interpretation of
  medical research.
\newblock \emph{Chest}, 158\penalty0 (1):\penalty0 S21--S28, 2020.

\bibitem[Finn et~al.(2017)Finn, Abbeel, and Levine]{finn2017model}
Chelsea Finn, Pieter Abbeel, and Sergey Levine.
\newblock Model-agnostic meta-learning for fast adaptation of deep networks.
\newblock \emph{arXiv preprint arXiv:1703.03400}, 2017.

\bibitem[Ganin et~al.(2016)Ganin, Ustinova, Ajakan, Germain, Larochelle,
  Laviolette, Marchand, and Lempitsky]{ganin2016domain}
Yaroslav Ganin, Evgeniya Ustinova, Hana Ajakan, Pascal Germain, Hugo
  Larochelle, Fran{\c{c}}ois Laviolette, Mario Marchand, and Victor Lempitsky.
\newblock Domain-adversarial training of neural networks.
\newblock \emph{The Journal of Machine Learning Research}, 17\penalty0
  (1):\penalty0 2096--2030, 2016.

\bibitem[Ghimire et~al.(2020)Ghimire, Kashyap, Wu, Karargyris, and
  Moradi]{ghimireLearningInvariantFeature2020}
Sandesh Ghimire, Satyananda Kashyap, Joy~T. Wu, Alexandros Karargyris, and
  Mehdi Moradi.
\newblock Learning {{Invariant Feature Representation}} to {{Improve
  Generalization}} across {{Chest X}}-ray {{Datasets}}.
\newblock \emph{arXiv:2008.04152 [cs, eess]}, August 2020.

\bibitem[Gulrajani and {Lopez-Paz}(2020)]{gulrajaniSearchLostDomain2020}
Ishaan Gulrajani and David {Lopez-Paz}.
\newblock In {{Search}} of {{Lost Domain Generalization}}.
\newblock \emph{arXiv:2007.01434 [cs, stat]}, July 2020.

\bibitem[Hardt et~al.(2016)Hardt, Price, and Srebro]{hardt2016equality}
Moritz Hardt, Eric Price, and Nathan Srebro.
\newblock Equality of opportunity in supervised learning, 2016.

\bibitem[{Heinze-Deml} et~al.(2018){Heinze-Deml}, Peters, and
  Meinshausen]{heinze-demlInvariantCausalPrediction2018}
Christina {Heinze-Deml}, Jonas Peters, and Nicolai Meinshausen.
\newblock Invariant {{Causal Prediction}} for {{Nonlinear Models}}.
\newblock \emph{Journal of Causal Inference}, 6\penalty0 (2), September 2018.
\newblock \doi{10.1515/jci-2017-0016}.

\bibitem[Hendrycks et~al.(2020)Hendrycks, Basart, Mu, Kadavath, Wang, Dorundo,
  Desai, Zhu, Parajuli, Guo, et~al.]{hendrycks2020many}
Dan Hendrycks, Steven Basart, Norman Mu, Saurav Kadavath, Frank Wang, Evan
  Dorundo, Rahul Desai, Tyler Zhu, Samyak Parajuli, Mike Guo, et~al.
\newblock The many faces of robustness: A critical analysis of
  out-of-distribution generalization.
\newblock \emph{arXiv preprint arXiv:2006.16241}, 2020.

\bibitem[Huang et~al.(2017)Huang, Liu, Van Der~Maaten, and
  Weinberger]{huang2017densely}
Gao Huang, Zhuang Liu, Laurens Van Der~Maaten, and Kilian~Q Weinberger.
\newblock Densely connected convolutional networks.
\newblock In \emph{Proceedings of the IEEE conference on computer vision and
  pattern recognition}, pages 4700--4708, 2017.

\bibitem[Irvin et~al.(2019)Irvin, Rajpurkar, Ko, Yu, Ciurea-Ilcus, Chute,
  Marklund, Haghgoo, Ball, Shpanskaya, Seekins, Mong, Halabi, Sandberg, Jones,
  Larson, Langlotz, Patel, Lungren, and Ng]{irvin_chexpert:_2019}
Jeremy Irvin, Pranav Rajpurkar, Michael Ko, Yifan Yu, Silviana Ciurea-Ilcus,
  Chris Chute, Henrik Marklund, Behzad Haghgoo, Robyn Ball, Katie Shpanskaya,
  Jayne Seekins, David~A. Mong, Safwan~S. Halabi, Jesse~K. Sandberg, Ricky
  Jones, David~B. Larson, Curtis~P. Langlotz, Bhavik~N. Patel, Matthew~P.
  Lungren, and Andrew~Y. Ng.
\newblock {CheXpert}: {A} {Large} {Chest} {Radiograph} {Dataset} with
  {Uncertainty} {Labels} and {Expert} {Comparison}.
\newblock \emph{arXiv:1901.07031 [cs, eess]}, January 2019.
\newblock arXiv: 1901.07031.

\bibitem[Johnson et~al.(2019)Johnson, Pollard, Berkowitz, Greenbaum, Lungren,
  Deng, Mark, and Horng]{johnson_mimic-cxr:_2019}
Alistair E.~W. Johnson, Tom~J. Pollard, Seth~J. Berkowitz, Nathaniel~R.
  Greenbaum, Matthew~P. Lungren, Chih-ying Deng, Roger~G. Mark, and Steven
  Horng.
\newblock {MIMIC}-{CXR}: {A} large publicly available database of labeled chest
  radiographs.
\newblock \emph{arXiv:1901.07042 [cs, eess]}, January 2019.

\bibitem[Johnson et~al.(2016)Johnson, Pollard, Shen, Li-Wei, Feng, Ghassemi,
  Moody, Szolovits, Celi, and Mark]{johnson2016mimic}
Alistair~EW Johnson, Tom~J Pollard, Lu~Shen, H~Lehman Li-Wei, Mengling Feng,
  Mohammad Ghassemi, Benjamin Moody, Peter Szolovits, Leo~Anthony Celi, and
  Roger~G Mark.
\newblock Mimic-iii, a freely accessible critical care database.
\newblock \emph{Scientific data}, 3\penalty0 (1):\penalty0 1--9, 2016.

\bibitem[Kamath et~al.(2021)Kamath, Tangella, Sutherland, and
  Srebro]{kamath2021does}
Pritish Kamath, Akilesh Tangella, Danica~J Sutherland, and Nathan Srebro.
\newblock Does invariant risk minimization capture invariance?
\newblock \emph{arXiv preprint arXiv:2101.01134}, 2021.

\bibitem[Kearns et~al.(2019)Kearns, Neel, Roth, and
  Wu]{kearns2019subgroupfairness}
Michael Kearns, Seth Neel, Aaron Roth, and Zhiwei~Steven Wu.
\newblock An empirical study of rich subgroup fairness for machine learning.
\newblock In \emph{Proceedings of the Conference on Fairness, Accountability,
  and Transparency}, FAT* '19, page 100–109, New York, NY, USA, 2019.
  Association for Computing Machinery.
\newblock ISBN 9781450361255.
\newblock \doi{10.1145/3287560.3287592}.
\newblock URL \url{https://doi.org/10.1145/3287560.3287592}.

\bibitem[Kelly et~al.(2019)Kelly, Karthikesalingam, Suleyman, Corrado, and
  King]{kellyKeyChallengesDelivering2019}
Christopher~J. Kelly, Alan Karthikesalingam, Mustafa Suleyman, Greg Corrado,
  and Dominic King.
\newblock Key challenges for delivering clinical impact with artificial
  intelligence.
\newblock \emph{BMC Medicine}, 17\penalty0 (1):\penalty0 195, October 2019.
\newblock ISSN 1741-7015.
\newblock \doi{10.1186/s12916-019-1426-2}.

\bibitem[Khani and Liang(2020)]{khani2020removing}
Fereshte Khani and Percy Liang.
\newblock Removing spurious features can hurt accuracy and affect groups
  disproportionately.
\newblock \emph{arXiv preprint arXiv:2012.04104}, 2020.

\bibitem[Koh et~al.(2020)Koh, Sagawa, Marklund, Xie, Zhang, Balsubramani, Hu,
  Yasunaga, Phillips, Beery, Leskovec, Kundaje, Pierson, Levine, Finn, and
  Liang]{kohWILDSBenchmarkIntheWild2020a}
Pang~Wei Koh, Shiori Sagawa, Henrik Marklund, Sang~Michael Xie, Marvin Zhang,
  Akshay Balsubramani, Weihua Hu, Michihiro Yasunaga, Richard~Lanas Phillips,
  Sara Beery, Jure Leskovec, Anshul Kundaje, Emma Pierson, Sergey Levine,
  Chelsea Finn, and Percy Liang.
\newblock {{WILDS}}: {{A Benchmark}} of in-the-{{Wild Distribution Shifts}}.
\newblock \emph{arXiv:2012.07421 [cs]}, December 2020.

\bibitem[Koyama and
  Yamaguchi(2020)]{koyamaOutofDistributionGeneralizationMaximal2020}
Masanori Koyama and Shoichiro Yamaguchi.
\newblock Out-of-{{Distribution Generalization}} with {{Maximal Invariant
  Predictor}}.
\newblock \emph{arXiv:2008.01883 [cs, stat]}, August 2020.

\bibitem[Krueger et~al.(2020)Krueger, Caballero, Jacobsen, Zhang, Binas, Priol,
  and Courville]{kruegerOutofDistributionGeneralizationRisk2020}
David Krueger, Ethan Caballero, Joern-Henrik Jacobsen, Amy Zhang, Jonathan
  Binas, Remi~Le Priol, and Aaron Courville.
\newblock Out-of-{{Distribution Generalization}} via {{Risk Extrapolation}}
  ({{REx}}).
\newblock \emph{arXiv:2003.00688 [cs, stat]}, March 2020.

\bibitem[Li et~al.(2017)Li, Yang, Song, and
  Hospedales]{liDeeperBroaderArtier2017}
Da~Li, Yongxin Yang, Yi-Zhe Song, and Timothy~M. Hospedales.
\newblock Deeper, {{Broader}} and {{Artier Domain Generalization}}.
\newblock \emph{arXiv:1710.03077 [cs]}, October 2017.

\bibitem[Li et~al.(2018{\natexlab{a}})Li, Yang, Song, and
  Hospedales]{li2018learning}
Da~Li, Yongxin Yang, Yi-Zhe Song, and Timothy Hospedales.
\newblock Learning to generalize: Meta-learning for domain generalization.
\newblock In \emph{Proceedings of the AAAI Conference on Artificial
  Intelligence}, volume~32, 2018{\natexlab{a}}.

\bibitem[Li et~al.(2018{\natexlab{b}})Li, Jialin~Pan, Wang, and
  Kot]{li2018domain}
Haoliang Li, Sinno Jialin~Pan, Shiqi Wang, and Alex~C Kot.
\newblock Domain generalization with adversarial feature learning.
\newblock In \emph{Proceedings of the IEEE Conference on Computer Vision and
  Pattern Recognition}, pages 5400--5409, 2018{\natexlab{b}}.

\bibitem[M{\aa}rtensson et~al.(2020)M{\aa}rtensson, Ferreira, Granberg,
  Cavallin, Oppedal, Padovani, Rektorova, Bonanni, Pardini, Kramberger, Taylor,
  Hort, Sn{\ae}dal, Kulisevsky, Blanc, Antonini, Mecocci, Vellas, Tsolaki,
  K{\l}oszewska, Soininen, Lovestone, Simmons, Aarsland, and
  Westman]{martenssonReliabilityDeepLearning2020}
Gustav M{\aa}rtensson, Daniel Ferreira, Tobias Granberg, Lena Cavallin, Ketil
  Oppedal, Alessandro Padovani, Irena Rektorova, Laura Bonanni, Matteo Pardini,
  Milica~G Kramberger, John-Paul Taylor, Jakub Hort, J{\'o}n Sn{\ae}dal, Jaime
  Kulisevsky, Frederic Blanc, Angelo Antonini, Patrizia Mecocci, Bruno Vellas,
  Magda Tsolaki, Iwona K{\l}oszewska, Hilkka Soininen, Simon Lovestone, Andrew
  Simmons, Dag Aarsland, and Eric Westman.
\newblock The reliability of a deep learning model in clinical
  out-of-distribution {{MRI}} data: {{A}} multicohort study.
\newblock \emph{Medical Image Analysis}, 66:\penalty0 101714, December 2020.
\newblock ISSN 1361-8415.
\newblock \doi{10.1016/j.media.2020.101714}.

\bibitem[Menon and Williamson(2018)]{menon2018cost}
Aditya~Krishna Menon and Robert~C Williamson.
\newblock The cost of fairness in binary classification.
\newblock In \emph{Conference on Fairness, Accountability and Transparency},
  pages 107--118. PMLR, 2018.

\bibitem[Moyer et~al.(2018)Moyer, Gao, Brekelmans, Galstyan, and
  Ver~Steeg]{moyer2018invariant}
Daniel Moyer, Shuyang Gao, Rob Brekelmans, Aram Galstyan, and Greg Ver~Steeg.
\newblock Invariant representations without adversarial training.
\newblock \emph{Advances in Neural Information Processing Systems},
  31:\penalty0 9084--9093, 2018.

\bibitem[Nestor et~al.(2019)Nestor, McDermott, Boag, Berner, Naumann, Hughes,
  Goldenberg, and Ghassemi]{nestorFeatureRobustnessNonstationary2019}
Bret Nestor, Matthew B.~A. McDermott, Willie Boag, Gabriela Berner, Tristan
  Naumann, Michael~C. Hughes, Anna Goldenberg, and Marzyeh Ghassemi.
\newblock Feature {{Robustness}} in {{Non}}-stationary {{Health Records}}:
  {{Caveats}} to {{Deployable Model Performance}} in {{Common Clinical Machine
  Learning Tasks}}.
\newblock \emph{arXiv:1908.00690 [cs, stat]}, August 2019.

\bibitem[Nordon et~al.(2019)Nordon, Koren, Shalev, Kimelfeld, Shalit, and
  Radinsky]{nordon2019building}
Galia Nordon, Gideon Koren, Varda Shalev, Benny Kimelfeld, Uri Shalit, and Kira
  Radinsky.
\newblock Building causal graphs from medical literature and electronic medical
  records.
\newblock In \emph{Proceedings of the AAAI Conference on Artificial
  Intelligence}, volume~33, pages 1102--1109, 2019.

\bibitem[Oakden-Rayner et~al.(2020)Oakden-Rayner, Dunnmon, Carneiro, and
  R{\'e}]{oakden2020hidden}
Luke Oakden-Rayner, Jared Dunnmon, Gustavo Carneiro, and Christopher R{\'e}.
\newblock Hidden stratification causes clinically meaningful failures in
  machine learning for medical imaging.
\newblock In \emph{Proceedings of the ACM conference on health, inference, and
  learning}, pages 151--159, 2020.

\bibitem[Pace et~al.(2020)Pace, Russo, and
  Shanahan]{paceLearningDiverseRepresentations2020}
Daniel Pace, Alessandra Russo, and Murray Shanahan.
\newblock Learning {{Diverse Representations}} for {{Fast Adaptation}} to
  {{Distribution Shift}}.
\newblock \emph{arXiv:2006.07119 [cs, stat]}, June 2020.

\bibitem[{Pan} and {Yang}(2010)]{Pan2009}
S.~J. {Pan} and Q.~{Yang}.
\newblock A survey on transfer learning.
\newblock \emph{IEEE Transactions on Knowledge and Data Engineering},
  22\penalty0 (10):\penalty0 1345--1359, 2010.

\bibitem[Perone et~al.(2019)Perone, Ballester, Barros, and
  Cohen-Adad]{perone2019unsupervised}
Christian~S Perone, Pedro Ballester, Rodrigo~C Barros, and Julien Cohen-Adad.
\newblock Unsupervised domain adaptation for medical imaging segmentation with
  self-ensembling.
\newblock \emph{NeuroImage}, 194:\penalty0 1--11, 2019.

\bibitem[Peters et~al.(2015)Peters, B{\"u}hlmann, and
  Meinshausen]{petersCausalInferenceUsing2015}
Jonas Peters, Peter B{\"u}hlmann, and Nicolai Meinshausen.
\newblock Causal inference using invariant prediction: Identification and
  confidence intervals.
\newblock \emph{arXiv:1501.01332 [stat]}, November 2015.

\bibitem[Pfohl et~al.(2021)Pfohl, Foryciarz, and Shah]{pfohl2021empirical}
Stephen~R Pfohl, Agata Foryciarz, and Nigam~H Shah.
\newblock An empirical characterization of fair machine learning for clinical
  risk prediction.
\newblock \emph{Journal of biomedical informatics}, 113:\penalty0 103621, 2021.

\bibitem[Piratla et~al.(2020)Piratla, Netrapalli, and
  Sarawagi]{piratlaEfficientDomainGeneralization2020}
Vihari Piratla, Praneeth Netrapalli, and Sunita Sarawagi.
\newblock Efficient {{Domain Generalization}} via {{Common}}-{{Specific
  Low}}-{{Rank Decomposition}}.
\newblock \emph{arXiv:2003.12815 [cs, stat]}, April 2020.

\bibitem[Pollard et~al.(2018)Pollard, Johnson, Raffa, Celi, Mark, and
  Badawi]{pollard2018eicu}
Tom~J Pollard, Alistair~EW Johnson, Jesse~D Raffa, Leo~A Celi, Roger~G Mark,
  and Omar Badawi.
\newblock The eicu collaborative research database, a freely available
  multi-center database for critical care research.
\newblock \emph{Scientific data}, 5:\penalty0 180178, 2018.

\bibitem[Pooch et~al.(2020)Pooch, Ballester, and Barros]{poochCanWeTrust2020}
Eduardo H.~P. Pooch, Pedro~L. Ballester, and Rodrigo~C. Barros.
\newblock Can we trust deep learning models diagnosis? {{The}} impact of domain
  shift in chest radiograph classification.
\newblock \emph{arXiv:1909.01940 [cs, eess, stat]}, June 2020.

\bibitem[Raghu et~al.(2019)Raghu, Zhang, Kleinberg, and
  Bengio]{raghu_transfusion:_2019}
Maithra Raghu, Chiyuan Zhang, Jon Kleinberg, and Samy Bengio.
\newblock Transfusion: {Understanding} {Transfer} {Learning} for {Medical}
  {Imaging}.
\newblock In \emph{Proceedings of the 33rd {International} {Conference} on
  {Neural} {Information} {Processing} {Systems}}, 2019.

\bibitem[Rajkomar et~al.(2018)Rajkomar, Oren, Chen, Dai, Hajaj, Hardt, Liu,
  Liu, Marcus, Sun, et~al.]{rajkomar2018scalable}
Alvin Rajkomar, Eyal Oren, Kai Chen, Andrew~M Dai, Nissan Hajaj, Michaela
  Hardt, Peter~J Liu, Xiaobing Liu, Jake Marcus, Mimi Sun, et~al.
\newblock Scalable and accurate deep learning with electronic health records.
\newblock \emph{NPJ Digital Medicine}, 1\penalty0 (1):\penalty0 18, 2018.

\bibitem[Rajpurkar et~al.(2018)Rajpurkar, Irvin, Ball, Zhu, Yang, Mehta, Duan,
  Ding, Bagul, Langlotz, Patel, Yeom, Shpanskaya, Blankenberg, Seekins,
  Amrhein, Mong, Halabi, Zucker, Ng, and Lungren]{rajpurkar_deep_2018}
Pranav Rajpurkar, Jeremy Irvin, Robyn~L. Ball, Kaylie Zhu, Brandon Yang,
  Hershel Mehta, Tony Duan, Daisy Ding, Aarti Bagul, Curtis~P. Langlotz,
  Bhavik~N. Patel, Kristen~W. Yeom, Katie Shpanskaya, Francis~G. Blankenberg,
  Jayne Seekins, Timothy~J. Amrhein, David~A. Mong, Safwan~S. Halabi, Evan~J.
  Zucker, Andrew~Y. Ng, and Matthew~P. Lungren.
\newblock Deep learning for chest radiograph diagnosis: {A} retrospective
  comparison of the {CheXNeXt} algorithm to practicing radiologists.
\newblock \emph{PLOS Medicine}, 15\penalty0 (11):\penalty0 e1002686, November
  2018.
\newblock ISSN 1549-1676.
\newblock \doi{10.1371/journal.pmed.1002686}.
\newblock URL \url{http://dx.plos.org/10.1371/journal.pmed.1002686}.

\bibitem[Rosenfeld et~al.(2020)Rosenfeld, Ravikumar, and
  Risteski]{rosenfeldRisksInvariantRisk2020}
Elan Rosenfeld, Pradeep Ravikumar, and Andrej Risteski.
\newblock The {{Risks}} of {{Invariant Risk Minimization}}.
\newblock \emph{arXiv:2010.05761 [cs, stat]}, October 2020.

\bibitem[Rothenh{\"a}usler et~al.(2019)Rothenh{\"a}usler, Meinshausen,
  B{\"u}hlmann, and Peters]{rothenhauslerAnchorRegressionHeterogeneous2019}
Dominik Rothenh{\"a}usler, Nicolai Meinshausen, Peter B{\"u}hlmann, and Jonas
  Peters.
\newblock Anchor regression: Heterogeneous data meets causality.
\newblock \emph{arXiv:1801.06229 [stat]}, June 2019.

\bibitem[Sagawa et~al.(2019)Sagawa, Koh, Hashimoto, and
  Liang]{sagawa2019distributionally}
Shiori Sagawa, Pang~Wei Koh, Tatsunori~B Hashimoto, and Percy Liang.
\newblock Distributionally robust neural networks for group shifts: On the
  importance of regularization for worst-case generalization.
\newblock \emph{arXiv preprint arXiv:1911.08731}, 2019.

\bibitem[Seyyed-Kalantari et~al.(2020)Seyyed-Kalantari, Liu, McDermott, and
  Marzyeh]{CheXclusion_2020}
Laleh Seyyed-Kalantari, Guanxiong Liu, Matthew McDermott, and Ghassemi Marzyeh.
\newblock Chexclusion: Fairness gaps in deep chest x-ray classifiers.
\newblock \emph{arXiv preprint arXiv:2003.00827}, 2020.

\bibitem[Sharifi-Malvajerdi et~al.(2019)Sharifi-Malvajerdi, Kearns, and
  Roth]{sharifi2019individualfairness}
Saeed Sharifi-Malvajerdi, Michael Kearns, and Aaron Roth.
\newblock Average individual fairness: Algorithms, generalization and
  experiments.
\newblock In H.~Wallach, H.~Larochelle, A.~Beygelzimer, F.~d\textquotesingle
  Alch\'{e}-Buc, E.~Fox, and R.~Garnett, editors, \emph{Advances in Neural
  Information Processing Systems}, volume~32, pages 8242--8251. Curran
  Associates, Inc., 2019.
\newblock URL
  \url{https://proceedings.neurips.cc/paper/2019/file/0e1feae55e360ff05fef58199b3fa521-Paper.pdf}.

\bibitem[Sheikhalishahi et~al.(2020)Sheikhalishahi, Balaraman, and
  Osmani]{sheikhalishahi2020benchmarking}
Seyedmostafa Sheikhalishahi, Vevake Balaraman, and Venet Osmani.
\newblock Benchmarking machine learning models on multi-centre eicu critical
  care dataset.
\newblock \emph{Plos one}, 15\penalty0 (7):\penalty0 e0235424, 2020.

\bibitem[Stacke et~al.(2020)Stacke, Eilertsen, Unger, and
  Lundstrom]{stackeMeasuringDomainShift2020}
K.~Stacke, G.~Eilertsen, J.~Unger, and C.~Lundstrom.
\newblock Measuring {{Domain Shift}} for {{Deep Learning}} in
  {{Histopathology}}.
\newblock \emph{IEEE Journal of Biomedical and Health Informatics}, pages 1--1,
  2020.
\newblock ISSN 2168-2208.
\newblock \doi{10.1109/JBHI.2020.3032060}.

\bibitem[Stacke et~al.(2019)Stacke, Eilertsen, Unger, and
  Lundstr{\"o}m]{stackeCloserLookDomain2019}
Karin Stacke, Gabriel Eilertsen, Jonas Unger, and Claes Lundstr{\"o}m.
\newblock A {{Closer Look}} at {{Domain Shift}} for {{Deep Learning}} in
  {{Histopathology}}.
\newblock \emph{arXiv:1909.11575 [cs]}, September 2019.

\bibitem[Subbaswamy and
  Saria(2020)]{subbaswamyDevelopmentDeploymentDataset2020}
Adarsh Subbaswamy and Suchi Saria.
\newblock From development to deployment: Dataset shift, causality, and
  shift-stable models in health {{AI}}.
\newblock \emph{Biostatistics}, 21\penalty0 (2):\penalty0 345--352, April 2020.
\newblock ISSN 1465-4644.
\newblock \doi{10.1093/biostatistics/kxz041}.

\bibitem[Sun and Saenko(2016)]{sun2016deep}
Baochen Sun and Kate Saenko.
\newblock Deep coral: Correlation alignment for deep domain adaptation.
\newblock In \emph{European conference on computer vision}, pages 443--450.
  Springer, 2016.

\bibitem[Thagaard et~al.(2020)Thagaard, Hauberg, {van der Vegt}, Ebstrup,
  Hansen, and Dahl]{thagaardCanYouTrust2020}
Jeppe Thagaard, S{\o}ren Hauberg, Bert {van der Vegt}, Thomas Ebstrup, Johan~D.
  Hansen, and Anders~B. Dahl.
\newblock Can {{You Trust Predictive Uncertainty Under Real Dataset Shifts}} in
  {{Digital Pathology}}?
\newblock In Anne~L. Martel, Purang Abolmaesumi, Danail Stoyanov, Diana Mateus,
  Maria~A. Zuluaga, S.~Kevin Zhou, Daniel Racoceanu, and Leo Joskowicz,
  editors, \emph{Medical {{Image Computing}} and {{Computer Assisted
  Intervention}} \textendash{} {{MICCAI}} 2020}, volume 12261, pages 824--833.
  {Springer International Publishing}, {Cham}, 2020.
\newblock ISBN 978-3-030-59709-2 978-3-030-59710-8.
\newblock \doi{10.1007/978-3-030-59710-8_80}.

\bibitem[Toma{\v{s}}ev et~al.(2019)Toma{\v{s}}ev, Glorot, Rae, Zielinski,
  Askham, Saraiva, Mottram, Meyer, Ravuri, Protsyuk,
  et~al.]{tomavsev2019clinically}
Nenad Toma{\v{s}}ev, Xavier Glorot, Jack~W Rae, Michal Zielinski, Harry Askham,
  Andre Saraiva, Anne Mottram, Clemens Meyer, Suman Ravuri, Ivan Protsyuk,
  et~al.
\newblock A clinically applicable approach to continuous prediction of future
  acute kidney injury.
\newblock \emph{Nature}, 572\penalty0 (7767):\penalty0 116--119, 2019.

\bibitem[Vapnik(1992)]{vapnik1992principles}
Vladimir Vapnik.
\newblock Principles of risk minimization for learning theory.
\newblock In \emph{Advances in neural information processing systems}, pages
  831--838, 1992.

\bibitem[Wang et~al.(2017)Wang, Peng, Lu, Lu, Bagheri, and
  Summers]{wang_chestx-ray8:_2017}
Xiaosong Wang, Yifan Peng, Le~Lu, Zhiyong Lu, Mohammadhadi Bagheri, and
  Ronald~M. Summers.
\newblock {ChestX}-ray8: {Hospital}-{Scale} {Chest} {X}-{Ray} {Database} and
  {Benchmarks} on {Weakly}-{Supervised} {Classification} and {Localization} of
  {Common} {Thorax} {Diseases}.
\newblock In \emph{Computer Vision and Pattern Recognition (CVPR) 2017}, pages
  2097--2106. IEEE, 2017.
\newblock URL
  \url{http://openaccess.thecvf.com/content_cvpr_2017/html/Wang_ChestX-ray8_Hospital-Scale_Chest_CVPR_2017_paper.html}.

\bibitem[Wu et~al.(2019)Wu, Kobayashi, Ding, Cheng, and
  Ghassemi]{wu2019modeling}
Denny Wu, Hirofumi Kobayashi, Charles Ding, Lei Cheng, and Keisuke Goda~Marzyeh
  Ghassemi.
\newblock Modeling the biological pathology continuum with hsic-regularized
  wasserstein auto-encoders.
\newblock \emph{arXiv preprint arXiv:1901.06618}, 2019.

\bibitem[Xie et~al.(2020)Xie, Chen, Liu, and
  Li]{xieRiskVariancePenalization2020}
Chuanlong Xie, Fei Chen, Yue Liu, and Zhenguo Li.
\newblock Risk {{Variance Penalization}}: {{From Distributional Robustness}} to
  {{Causality}}.
\newblock \emph{arXiv:2006.07544 [cs, stat]}, June 2020.

\bibitem[Yule(1912)]{yule1912methods}
G~Udny Yule.
\newblock On the methods of measuring association between two attributes.
\newblock \emph{Journal of the Royal Statistical Society}, 75\penalty0
  (6):\penalty0 579--652, 1912.

\bibitem[Zech et~al.(2018)Zech, Badgeley, Liu, Costa, Titano, and
  Oermann]{zechVariableGeneralizationPerformance2018}
John~R. Zech, Marcus~A. Badgeley, Manway Liu, Anthony~B. Costa, Joseph~J.
  Titano, and Eric~Karl Oermann.
\newblock Variable generalization performance of a deep learning model to
  detect pneumonia in chest radiographs: {{A}} cross-sectional study.
\newblock \emph{PLOS Medicine}, 15\penalty0 (11):\penalty0 e1002683, November
  2018.
\newblock ISSN 1549-1676.
\newblock \doi{10.1371/journal.pmed.1002683}.

\bibitem[Zhang et~al.(2019)Zhang, Wang, Yang, Sanford, Harmon, Turkbey, Roth,
  Myronenko, Xu, and Xu]{zhangWhenUnseenDomain2019}
Ling Zhang, Xiaosong Wang, Dong Yang, Thomas Sanford, Stephanie Harmon, Baris
  Turkbey, Holger Roth, Andriy Myronenko, Daguang Xu, and Ziyue Xu.
\newblock When {{Unseen Domain Generalization}} is {{Unnecessary}}?
  {{Rethinking Data Augmentation}}.
\newblock \emph{arXiv:1906.03347 [cs, eess]}, June 2019.

\bibitem[Zhang et~al.(2020)Zhang, Wei, Wu, Zhao, Niu, Huang, and
  Tan]{zhang2020collaborative}
Yifan Zhang, Ying Wei, Qingyao Wu, Peilin Zhao, Shuaicheng Niu, Junzhou Huang,
  and Mingkui Tan.
\newblock Collaborative unsupervised domain adaptation for medical image
  diagnosis.
\newblock \emph{IEEE Transactions on Image Processing}, 29:\penalty0
  7834--7844, 2020.

\end{thebibliography}

\clearpage
\onecolumn
\begin{appendices}
\setcounter{table}{0}
\renewcommand{\thetable}{A\arabic{table}}

\setcounter{figure}{0}
\renewcommand{\thefigure}{A\arabic{figure}}

\section{Dataset Statistics}
\label{sec_app:dataset}

\begin{table}[!h]
\caption{\label{tab:eicu_full_stats}Statistics for each region for the eICU in-hospital mortality prediction task}
\begin{tabular}{l|r|r|r|r|r}
\textbf{Region}              & \multicolumn{1}{r|}{\textbf{Midwest}} & \multicolumn{1}{r|}{\textbf{West}} & \multicolumn{1}{r|}{\textbf{Northeast}} & \multicolumn{1}{r|}{\textbf{Missing}} & \multicolumn{1}{r}{\textbf{South}} \\ \hline

\textbf{Assigned Split} & Train & Train & Train & Validation & Test \\ \hline

\textbf{\# Patients}         & 10,985                                & 4,527                              & 2,495                                   & 1,846                                 & 10,827                             \\
\textbf{\% Positive}         & 9.43\%                                & 14.42\%                            & 13.19\%                                 & 12.68\%                               & 11.74\%                            \\
\textbf{\# Unique Hospitals} & 69                                    & 41                                 & 13                                      & 25                                    & 52                                 \\
\textbf{Mean Age}            & 63.4                                  & 64.0                               & 63.9                                    & 65.3                                  & 63.8                               \\ \hline
\textbf{Male}                & 54.55\%                               & 55.75\%                            & 54.07\%                                 & 55.58\%                               & 54.12\%                            \\
\textbf{Female}              & 45.45\%                               & 44.25\%                            & 45.93\%                                 & 44.42\%                               & 45.88\%                            \\ \hline
\textbf{African American}    & 8.33\%                                & 2.64\%                             & 2.40\%                                  & 5.99\%                                & 20.80\%                            \\
\textbf{Asian}               & 1.01\%                                & 3.70\%                             & 0.52\%                                  & 4.23\%                                & 1.18\%                             \\
\textbf{Caucasian}           & 83.68\%                               & 78.43\%                            & 92.34\%                                 & 73.68\%                               & 69.43\%                            \\
\textbf{Hispanic}            & 1.36\%                                & 6.21\%                             & 0.80\%                                  & 8.90\%                                & 4.72\%                             \\
\textbf{Native American}     & 0.52\%                                & 1.04\%                             & 0.12\%                                  & 0.49\%                                & 0.33\%                             \\
\textbf{Other/Unknown}       & 5.10\%                                & 7.98\%                             & 3.81\%                                  & 6.70\%                                & 3.54\%                            
\end{tabular}
\end{table}

\begin{table}[h!]
\caption{Features used for the eICU mortality prediction task. \label{tab:eicu_features}}
\begin{tabular}{ll|l|l}
\multicolumn{2}{l|}{\textbf{Time-Series}}                          & \multicolumn{2}{l}{\textbf{Static}}       \\ \hline
\multicolumn{1}{l|}{\textbf{Continuous}}   & \textbf{Categorical} & \textbf{Continuous} & \textbf{Categorical} \\  \hline
\multicolumn{1}{l|}{Heart Rate}            & GCS Total            & Admission Height    & Admission Diagnosis  \\
\multicolumn{1}{l|}{MAP}                   & Eyes                 & Admission Weight    & Gender              \\
\multicolumn{1}{l|}{Invasive BP Diastolic} & Motor                & Age                 &                      \\
\multicolumn{1}{l|}{Invasive BP Systolic}  & Verbal               &                     &                      \\
\multicolumn{1}{l|}{O2 Saturation}         &                      &                     &                      \\
\multicolumn{1}{l|}{Respiratory Rate}      &                      &                     &                      \\
\multicolumn{1}{l|}{Temperature}           &                      &                     &                      \\
\multicolumn{1}{l|}{glucose}               &                      &                     &                      \\
\multicolumn{1}{l|}{FiO2}                  &                      &                     &                      \\
\multicolumn{1}{l|}{pH}                    &                      &                     &                     
\end{tabular}
\end{table}

\begin{table}[!h]
\caption{\label{tab:cxr_full_stats} Summary statistics for the four chest X-ray datasets. Note that we use only frontal images for our experiments. Demographics are shown for patients in the whole dataset, while label distributions shown are only for frontal images.}
\begin{tabular}{l|r|r|r|r}
                       & \multicolumn{1}{l|}{\textbf{MIMIC-CXR}} & \multicolumn{1}{l|}{\textbf{CheXpert}} & \multicolumn{1}{l|}{\textbf{Chest-Xray8}} & \multicolumn{1}{l}{\textbf{PadChest}}                                                                        \\ \hline
\textbf{Assigned Split}          & Train                                   & Train                                  & Validation                                & Test                                                        \\ \hline
\textbf{Location}      & Boston                                  & Stanford                               & Bethesda                                  & \begin{tabular}[c]{@{}r@{}}Alicante \\ (Spain)\end{tabular} \\ \hline
\textbf{\# Images}     & 371,858                                 & 223,648                                & 112,120                                   & 144,639                                                     \\
\textbf{\# Patients}   & 65,079                                  & 64,740                                 & 30,805                                    & 64,874                                                      \\
\textbf{\# Frontal}    & 249,995                                 & 191,229                                & 112,120                                   & 99,934                                                      \\
\textbf{\# Lateral}    & 121,863                                 & 32,419                                 & 0                                         & 44,705                                                      \\ \hline
\textbf{Male}          & 52.17\%                                 & 59.36\%                                & 56.49\%                                   & 49.58\%                                                     \\ 
\textbf{Female}        & 47.83\%                                 & 40.64\%                                & 43.51\%                                   & 50.41\%                                                     \\\hline

\textbf{0-20}          & 2.20\%                                  & 0.87\%                                 & 6.09\%                                    & 4.06\%                                                      \\
\textbf{20-40}         & 19.51\%                                 & 13.18\%                                & 25.96\%                                   & 8.82\%                                                      \\
\textbf{40-60}         & 37.20\%                                 & 31.00\%                                & 43.83\%                                   & 26.54\%                                                     \\
\textbf{60-80}         & 34.12\%                                 & 38.94\%                                & 23.11\%                                   & 37.95\%                                                     \\
\textbf{80-}           & 6.97\%                                  & 16.01\%                                & 1.01\%                                    & 22.64\%                                                     \\ \hline
\textbf{No Finding}    & 33.33\%                                 & 8.89\%                                 & 53.84\%                                   & 36.19\%                                                     \\
\textbf{Atelectasis}   & 19.98\%                                 & 15.58\%                                & 10.31\%                                   & 5.49\%                                                      \\
\textbf{Cardiomegaly}  & 19.70\%                                 & 12.26\%                                & 2.48\%                                    & 9.08\%                                                      \\
\textbf{Effusion}      & 23.64\%                                 & 40.25\%                                & 11.88\%                                   & 6.01\%                                                      \\
\textbf{Pneumonia}     & 7.37\%                                  & 2.45\%                                 & 1.28\%                                    & 4.90\%                                                      \\
\textbf{Pneumothorax}  & 4.67\%                                  & 9.26\%                                 & 4.73\%                                    & 0.35\%                                                      \\
\textbf{Consolidation} & 4.73\%                                  & 6.81\%                                 & 4.16\%                                    & 1.56\%                                                      \\
\textbf{Edema}         & 11.82\%                                 & 26.00\%                                & 2.05\%                                    & 1.20\%                                                     
\end{tabular}
\end{table}

\FloatBarrier
\vspace*{20mm}
\section{Colored MNIST}
\label{sec_app:cmnist}
\subsection{Data Generation Process}
Inspired by the work of \citet{choeEmpiricalStudyInvariant2020} to vary data generation parameters in Colored MNIST, we introduce the following Colored MNIST data generation process with three adjustable parameters.

\begin{enumerate}
     \item Randomly split the MNIST data into 2 training environments ($e_1$, $e_2$, each with $n = 25,000$), and one test environment ($e_{test}$, $n = 10,000$).
    \item Generate a binary label $\hat{y}_{obs}$ from the MNIST label $y_{num}$ by assigning $\hat{y}_{obs} = 0$ if $y_{num} \in \{0, 1, ..., 4\}$ and $\hat{y}_{obs} = 1$ otherwise.
    \item Flip $\hat{y}_{obs}$ with probability $\eta$ to obtain $y$.
    \item Define $\hat{x}_{ch} = y$. Flip $\hat{x}_{ch}$ to obtain $x_{ch}$ with probability $p_1$, $p_2$, and $p_{test} = 0.9$ for $e_1$, $e_2$ and $e_{test}$ respectively. Here $p_1 = \beta + \delta/2$, $p_2 = \beta - \delta/2$.
    \item Construct $X$ as $[X_{fig} \cdot (1 - x_{ch}),\  X_{fig} \cdot x_{ch}]$.
\end{enumerate}

\noindent The three parameters that we vary are the following:
\begin{itemize}
    \item $\eta$: the label corruption probability. Corresponds to 1 - strength of the invariant correlation. Baseline value of 0.25.
    
    \item $\beta$: Average color flip probability between the two training environments. Corresponds to 1 - strength of the spurious correlation. Baseline value of 0.15.
    
    \item $\delta$: Gap between two training environments. Baseline value of 0.1. 
\end{itemize}

\subsection{Experiments}
We vary each data generation parameter independently while keeping the others at their baseline values (which correspond to their settings in standard Colored MNIST). To model the data, we flatten the image and use a dense MLP with ReLU activations, consistent with prior work \cite{arjovskyInvariantRiskMinimization2019}. For each data generation setting and for each model, we use 20 iterations of random search over model hyperparameters (number of layers, number of units, dropout probability), algorithm hyperparameters (such as the penalty coefficient and the number of annealing iterations), and optimization hyperparameters (batch size and learning rate). 

For early stopping and model selection, we experiment with using model accuracy for the following two schemes:
\begin{itemize}
    \item Training Domains: We use the pooled validation sets from the two training environments.
    \item Test Domain: We use a validation split from the test environment. This is the only model selection scheme used in the prior work for most domain generalization methods \cite{arjovskyInvariantRiskMinimization2019, kruegerOutofDistributionGeneralizationRisk2020, koyamaOutofDistributionGeneralizationMaximal2020, ahuja2020invariant, xieRiskVariancePenalization2020}. Note that as there are only three environments, we do not allocate a validation environment for Colored MNIST.
\end{itemize}

\subsection{Results}
\begin{figure}[!h]
    \centering
    \includegraphics[width=1\textwidth]{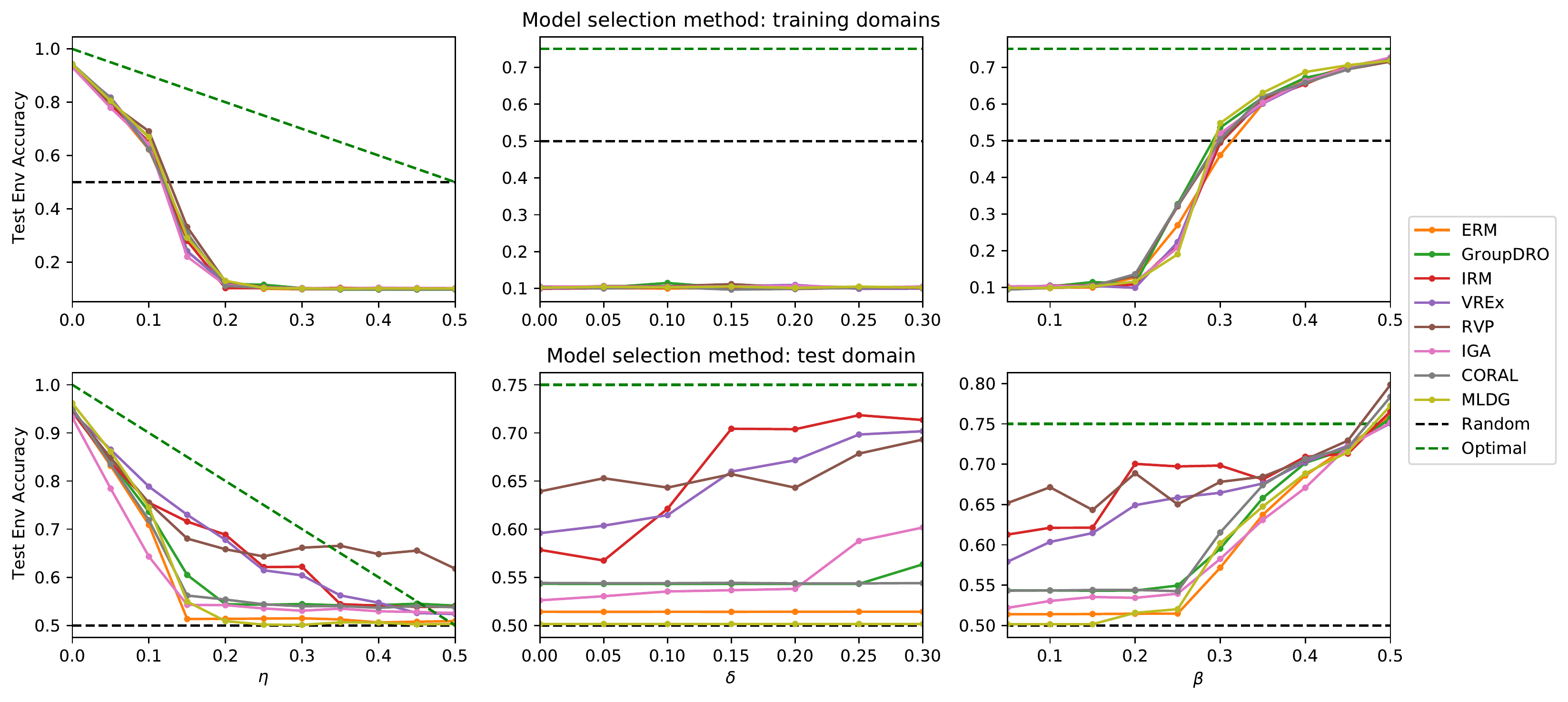}
    \caption{Test environment accuracy of various domain generalization methods on the Colored MNIST dataset.}
    \label{fig:cmnist_figure}
\end{figure}

First, we observe from Figure \ref{fig:cmnist_figure} that when the training domain is used for model selection, no method performs significantly better than ERM. The performance gains for domain generalization methods only appear when model selection is done directly on the test set. This is consistent with prior findings \cite{gulrajaniSearchLostDomain2020}. Though this is the setup used in the large majority of papers proposing domain generalization methods ~\cite{arjovskyInvariantRiskMinimization2019, kruegerOutofDistributionGeneralizationRisk2020, koyamaOutofDistributionGeneralizationMaximal2020, ahuja2020invariant, xieRiskVariancePenalization2020}, having test environment data for model selection is not realistic, and defeats much of the purpose of domain generalization. This is also a potential explanation for why domain generalization methods, which work well on Colored MNIST in the literature, do not work well in our clinical experiments (which do not use the test domain for model selection).

We also observe many intuitive correlations between the data generation parameters and the test environment accuracy. Increasing $\eta$, the data corruption probability, tends to decrease model performance, consistent with prior work \cite{choeEmpiricalStudyInvariant2020}. There is a pronounced increase in model performance as the training environment characteristics ($\beta$) move closer to the test environment ($0.9$). There is a limited increase in model performance with increasing environment diversity. Surprisingly, some models are able to achieve better performance than ERM even when the two training environments originate from the same distribution. We attribute this to the use of early stopping based on the validation metric in our experiments, as well as the variance in samples from minibatches drawn between the training environments.

Finally, we note the shift in reliance on the spurious versus invariant correlations that occur as we change certain data generation parameters. This is most pronounced when the training domains are used for model selection. For example, as we increase $\eta$ above $0.15$, the spurious correlation becomes stronger than the invariant correlation, and there is a marked drop in accuracy. Similarly, as we increase $\beta$ above $0.25$, the spurious correlation becomes weaker than the invariant correlation, and there is a decided increase in test domain accuracy.

\FloatBarrier

\clearpage
\section{Additional Results for Subsampling Shifts}
\label{sec_app:additional_dg}

\begin{table}[!h]
\caption{\label{tab:tpr_subsample} TPR gaps for \texttt{BiasSampUnobs} and \texttt{BiasSampObs}. We evaluate the test environment true positive rate gaps (M-F) in subsampling experiments with eICU and CXR datasets. We notice that observing the subsampled feature greatly increases the TPR disparity. Though there exist instances where domain generalization methods have lower disparity than ERM, the corresponding models also have lower AUROC.}
\resizebox{.99\textwidth}{!} {
\begin{tabular}{l|l|l|rr|rrrrrrrr}
\textbf{Dataset}                                                                 & \textbf{\begin{tabular}[c]{@{}l@{}}Selection\\ Method\end{tabular}} & \textbf{Observed} & \textbf{OracleID} & \textbf{OracleMerged} & \textbf{ERM}         & \textbf{GroupDRO}    & \textbf{IRM} & \textbf{VREx}        & \textbf{RVP}         & \textbf{IGA}         & \textbf{CORAL}       & \textbf{MLDG} \\ \hline
\multirow{4}{*}{\textbf{eICU}}                                                   & \multirow{2}{*}{\textbf{Training}}                                  & \textbf{No}       & -0.179±0.041      & 0.110±0.065           & 0.249±0.046          & 0.254±0.072          & 0.303±0.093  & 0.277±0.038          & 0.318±0.040          & 0.210±0.090 & 0.323±0.043          & 0.320±0.041   \\
                                                                                 &                                                                     & \textbf{Yes}      & -0.369±0.044      & 0.196±0.095           & 0.530±0.046          & 0.538±0.066 & 0.581±0.066  & 0.534±0.068          & 0.504±0.027 & 0.503±0.068          & 0.556±0.038          & 0.523±0.070   \\ \cline{2-13} 
                                                                                 & \multirow{2}{*}{\textbf{Validation}}                                & \textbf{No}       & -0.179±0.041      & 0.110±0.065           & 0.188±0.071          & 0.300±0.088          & 0.184±0.095  & 0.235±0.124          & 0.253±0.069 & 0.094±0.109          & 0.278±0.057          & 0.174±0.075   \\
                                                                                 &                                                                     & \textbf{Yes}      & -0.369±0.044      & 0.196±0.095           & 0.526±0.066          & 0.453±0.035          & 0.441±0.093  & 0.399±0.133 & 0.362±0.177          & 0.361±0.213          & 0.462±0.108          & 0.482±0.062   \\ \hline
\multirow{4}{*}{\textbf{\begin{tabular}[c]{@{}l@{}}CXR\\ (Binary)\end{tabular}}} & \multirow{2}{*}{\textbf{Training}}                                  & \textbf{No}       & 0.022±0.039       & 0.173±0.092           & 0.310±0.036          & 0.292±0.035          & 0.287±0.050  & 0.276±0.045          & 0.250±0.058          & 0.319±0.055          & 0.342±0.074 & 0.158±0.121   \\
                                                                                 &                                                                     & \textbf{Yes}      & 0.036±0.072       & 0.135±0.029           & 0.365±0.058 & 0.328±0.114          & 0.354±0.083  & 0.338±0.118          & 0.253±0.150          & 0.505±0.183          & 0.339±0.067          & 0.221±0.116   \\ \cline{2-13} 
                                                                                 & \multirow{2}{*}{\textbf{Validation}}                                & \textbf{No}       & 0.022±0.039       & 0.173±0.092           & 0.267±0.041          & 0.315±0.071          & 0.281±0.037  & 0.308±0.058 & 0.304±0.064          & 0.296±0.098          & 0.308±0.063          & 0.152±0.122   \\
                                                                                 &                                                                     & \textbf{Yes}      & 0.036±0.072       & 0.135±0.029           & 0.311±0.023          & 0.281±0.048          & 0.236±0.081  & 0.346±0.062          & 0.291±0.061          & 0.255±0.109 & 0.291±0.056          & 0.208±0.095  
\end{tabular}

}
\end{table}

\begin{table}[!h]
\caption{\label{tab:tnr_subsample} TNR gaps for \texttt{BiasSampUnobs} and \texttt{BiasSampObs}. We evaluate the test environment true negative rate gaps (M-F) in subsampling experiments with eICU and Chest X-ray (CXR) datasets. }
\resizebox{.99\textwidth}{!} {
\begin{tabular}{l|l|l|rr|rrrrrrrr}
\textbf{Dataset}                                                                 & \textbf{\begin{tabular}[c]{@{}l@{}}Selection\\ Method\end{tabular}} & \textbf{Observed} & \textbf{OracleID} & \textbf{OracleMerged} & \textbf{ERM}          & \textbf{GroupDRO}     & \textbf{IRM} & \textbf{VREx}         & \textbf{RVP}          & \textbf{IGA}          & \textbf{CORAL}        & \textbf{MLDG} \\ \hline
\multirow{4}{*}{\textbf{eICU}}                                                   & \multirow{2}{*}{\textbf{Training}}                                  & \textbf{No}       & 0.081±0.050       & -0.033±0.015          & -0.100±0.031          & -0.103±0.036          & -0.107±0.034 & -0.103±0.034          & -0.100±0.026          & -0.077±0.061 & -0.119±0.017          & -0.136±0.060  \\
                                                                                 &                                                                     & \textbf{Yes}      & 0.319±0.109       & -0.064±0.024          & -0.136±0.035          & -0.200±0.052 & -0.196±0.069 & -0.133±0.048          & -0.157±0.041 & -0.128±0.030          & -0.159±0.042          & -0.149±0.039  \\ \cline{2-13} 
                                                                                 & \multirow{2}{*}{\textbf{Validation}}                                & \textbf{No}       & 0.081±0.050       & -0.033±0.015          & -0.120±0.029          & -0.093±0.024          & -0.090±0.051 & -0.114±0.072          & -0.098±0.019 & -0.100±0.063          & -0.099±0.058          & -0.100±0.041  \\
                                                                                 &                                                                     & \textbf{Yes}      & 0.319±0.109       & -0.064±0.024          & -0.171±0.042          & -0.149±0.078          & -0.102±0.025 & -0.153±0.079 & -0.118±0.065          & -0.107±0.065          & -0.162±0.102          & -0.183±0.057  \\ \hline
\multirow{4}{*}{\textbf{\begin{tabular}[c]{@{}l@{}}CXR\\ (Binary)\end{tabular}}} & \multirow{2}{*}{\textbf{Training}}                                  & \textbf{No}       & -0.016±0.005      & -0.059±0.024          & -0.135±0.035          & -0.129±0.017          & -0.146±0.031 & -0.117±0.039          & -0.135±0.054          & -0.446±0.096          & -0.117±0.035 & -0.157±0.137  \\
                                                                                 &                                                                     & \textbf{Yes}      & -0.016±0.006      & -0.049±0.016          & -0.168±0.050 & -0.156±0.060          & -0.137±0.070 & -0.134±0.064          & -0.126±0.081          & -0.576±0.269          & -0.154±0.074          & -0.269±0.167  \\ \cline{2-13} 
                                                                                 & \multirow{2}{*}{\textbf{Validation}}                                & \textbf{No}       & -0.016±0.005      & -0.059±0.024          & -0.156±0.064          & -0.149±0.025          & -0.153±0.046 & -0.143±0.061 & -0.176±0.028          & -0.258±0.126          & -0.121±0.029          & -0.144±0.173  \\
                                                                                 &                                                                     & \textbf{Yes}      & -0.016±0.006      & -0.049±0.016          & -0.141±0.042          & -0.170±0.043          & -0.128±0.094 & -0.169±0.042          & -0.147±0.076          & -0.140±0.090 & -0.123±0.032          & -0.192±0.144 
\end{tabular}
}
\end{table}

\begin{table}[!h]
\caption{\label{tab:phi_subsample} Matthews correlation coefficient between the gender attribute and the predicted label for the subsampling augmentation on the test environment. A positive value indicates correlation between males and positive predictions (and between females and negative predictions), and a negative value indicates correlation between males and negative predictions (and females and positive predictions).}
\resizebox{.99\textwidth}{!} {
\begin{tabular}{l|l|l|rr|rrrrrrrr}
\textbf{Dataset}                                                                 & \textbf{\begin{tabular}[c]{@{}l@{}}Selection\\ Method\end{tabular}} & \textbf{Observed} & \textbf{OracleID} & \textbf{OracleMerged} & \textbf{ERM}         & \textbf{GroupDRO}    & \textbf{IRM} & \textbf{VREx}        & \textbf{RVP}          & \textbf{IGA}          & \textbf{CORAL}       & \textbf{MLDG} \\ \hline
\multirow{4}{*}{\textbf{eICU}}                                                   & \multirow{2}{*}{\textbf{Training}}                                  & \textbf{No}       & -0.314±0.025      & -0.161±0.033          & -0.023±0.020         & -0.023±0.033         & -0.008±0.042 & -0.017±0.021         & 0.007±0.027           & -0.037±0.033 & -0.003±0.029         & 0.017±0.038   \\
                                                                                 &                                                                     & \textbf{Yes}      & -0.496±0.046      & -0.091±0.043          & 0.104±0.023          & 0.126±0.037 & 0.138±0.057  & 0.109±0.036          & 0.103±0.013  & 0.107±0.035           & 0.127±0.023          & 0.103±0.022   \\ \cline{2-13} 
                                                                                 & \multirow{2}{*}{\textbf{Validation}}                                & \textbf{No}       & -0.314±0.025      & -0.161±0.033          & -0.044±0.028         & -0.014±0.030         & -0.054±0.021 & -0.034±0.068         & -0.026±0.033 & -0.054±0.060          & -0.034±0.046         & -0.067±0.056  \\
                                                                                 &                                                                     & \textbf{Yes}      & -0.496±0.046      & -0.091±0.043          & 0.108±0.027          & 0.084±0.026          & 0.050±0.040  & 0.068±0.076 & 0.029±0.094           & 0.061±0.039           & 0.073±0.062          & 0.102±0.031   \\ \hline
\multirow{4}{*}{\textbf{\begin{tabular}[c]{@{}l@{}}CXR\\ (Binary)\end{tabular}}} & \multirow{2}{*}{\textbf{Training}}                                  & \textbf{No}       & 0.031±0.012       & 0.108±0.040           & 0.246±0.035          & 0.241±0.017          & 0.260±0.028  & 0.232±0.033          & 0.244±0.055           & 0.444±0.092           & 0.236±0.039 & 0.216±0.125   \\
                                                                                 &                                                                     & \textbf{Yes}      & 0.032±0.010       & 0.094±0.016           & 0.281±0.046 & 0.274±0.055          & 0.251±0.066  & 0.251±0.062          & 0.217±0.128           & 0.612±0.236           & 0.273±0.070          & 0.328±0.114   \\ \cline{2-13} 
                                                                                 & \multirow{2}{*}{\textbf{Validation}}                                & \textbf{No}       & 0.031±0.012       & 0.108±0.040           & 0.255±0.050          & 0.259±0.019          & 0.260±0.033  & 0.231±0.072 & 0.274±0.030           & 0.294±0.106           & 0.233±0.037          & 0.177±0.156   \\
                                                                                 &                                                                     & \textbf{Yes}      & 0.032±0.010       & 0.094±0.016           & 0.245±0.036          & 0.266±0.038          & 0.219±0.104  & 0.277±0.035          & 0.245±0.058           & 0.200±0.091  & 0.234±0.023          & 0.240±0.121  
\end{tabular}

}
\end{table}

\end{appendices}

\end{document}